\documentclass{article}

\usepackage[margin=3cm, bottom=3cm, footskip=1.5cm]{geometry}

\usepackage{amsmath,amssymb,amsthm}
\usepackage{algorithm}
\usepackage{algorithmicx,algpseudocode}
\usepackage[subrefformat=parens]{subcaption}
\usepackage[bbgreekl]{mathbbol}
\usepackage{graphicx}
\usepackage{bm}
\usepackage{url}

\newcommand{\E}{\mathbb{E}} 
\newcommand{\PP}{\mathbb{P}} 

\title{Boltzmann machines for time-series}
\author{Takayuki Osogami\\\\IBM Research - Tokyo}
\date{{\tt osogami@jp.ibm.com}\\
\ \\
version 1.0.1}

\begin{document}

\maketitle

\begin{abstract}
We review Boltzmann machines extended for time-series.  These models
 often have recurrent structure, and back propagration through time
 (BPTT) is used to learn their parameters.  The per-step computational
 complexity of BPTT in online learning, however, grows linearly with
 respect to the length of preceding time-series ({\it i.e.}, learning rule is
 not local in time), which limits the applicability of BPTT in online
 learning.  We then review dynamic Boltzmann machines (DyBMs), whose
 learning rule is local in time.  DyBM's learning rule
 relates to spike-timing dependent plasticity (STDP), which has been
 postulated and experimentally confirmed for biological neural networks.
\end{abstract}

\section{Introduction}

The Boltzmann machine is a stochastic model for representing probability
distributions over binary patterns \cite{BMsurvey1}.  In this paper, we
review Boltzmann machines that have been studied as stochastic
(generative) models of time-series.  Such Boltzmann machines define
probability distributions over time-series of binary patterns.  They
can also be modified to deal with time-series of real-valued patterns,
similar to Boltzmann machines modified for real-valued patterns ({\it
e.g.}, Gaussian Boltzmann machines; see Section~\ref{sec:BM:Gaussian}
from \cite{BMsurvey1}).  We will follow the probabilistic
representations of \cite{BMsurvey1} for intuitive interpretations in
terms of probabilities.

In Section~\ref{sec:nonrecurrent}, we start with a Conditional
Restricted Boltzmann Machine (CRBM) \cite{CRBM}, which is a conditional Boltzmann
machine (Section~\ref{sec:BM:discriminative} from \cite{BMsurvey1}) that gives conditional probability of the next pattern given a
fixed number of preceding patterns.  A limitation of a CRBM is that it
can take into account only the dependency within a fixed horizon, and as
we increase the length of this horizon, the complexity of learning grows
accordingly.

To overcome this limitation of CRBMs, researchers have proposed
Boltzmann machines having recurrent structures, which we review in
Section~\ref{sec:recurrent}.  These include spiking Boltzmann machines \cite{HinBro99},
temporal restricted Boltzmann machines (TRBMs) \cite{TRBM}, recurrent temporal
restricted Boltzmann machines (RTRBMs) \cite{RTRBM}, and extensions of those models.
A standard approach to learning those models having recurrent structures
is back propagation through time (BPTT).

However, BPTT is undesirable when we learn time-series in an online
manner, where we update the parameters of a model every time a new
pattern arrives.  Such online learning is needed when we want to quickly
adapt to a changing environment or when we do not have sufficient memory
to store the time-series.  Unfortunately, the per-step computational
complexity of BPTT in online learning grows linearly with respect to the
length of preceding time-series.  This computational complexity
limits the applicability of BPTT to online learning.

In Section~\ref{sec:DyBM}, we review the dynamic Boltzmann machine (DyBM)
\cite{DyBM,RT0967} and its extensions.  The DyBM's per-step computational complexity in online
learning is independent of the length of preceding time-series.  We
discuss how the learning rule of the DyBM relates to spike-timing
dependent plasticity (STDP), which has been postulated and
experimentally confirmed for biological neural networks.

This survey paper is based on a personal note prepared for the third of
the four parts of a tutorial given at the 26th International Joint
Conference on Artificial Intelligence (IJCAI-17) held in Melbourne,
Australia on August 21, 2017. See a tutorial
webpage\footnote{https://researcher.watson.ibm.com/researcher/view\_group.php?id=7834}
for information about the tutorial.  A survey corresponding to the first
part of the tutorial (Boltzmann machines and energy-based models) can be
found in \cite{BMsurvey1}.  We follow the definitions and notations used
in \cite{BMsurvey1}.

\section{Learning energy-based models for time-series}

Consider a possibly multi-dimensional time-series:
\begin{align}
\mathbf{x} \equiv (\mathbf{x}^{[t]})_{t=0}^T,
\end{align}
where $\mathbf{x}^{[t]}$ denotes the binary pattern (vector) at time $t$.  We will use $\mathbf{x}^{[s,t]}$ to denote the time-series of the patterns from time $s$ to $t$.

A goal of learning time-series is to maximize the log-likelihood of a given time-series $\mathbf{x}$ (or a collection of multiple time-series) with respect to the distribution $\PP_\theta(\cdot)$ defined by a model under consideration, where we use $\theta$ to denote the set of the parameters of the model:
 \begin{align}
  f(\theta) & \equiv
\log \PP_\theta(\mathbf{x})
 = \sum_{t=0}^T \log \PP_\theta(\mathbf{x}^{[t]} \,|\, \mathbf{x}^{[0,t-1]}),
\label{eq:DyBM:LL}
 \end{align}
where $\PP_\theta(\mathbf{x}^{[t]} \,|\, \mathbf{x}^{[0,t-1]})$ denotes the conditional probability that the pattern at time $t$ is $\mathbf{x}^{[t]}$ given that the patterns up to time $t-1$ is $\mathbf{x}^{[0,t-1]}$.  Here, $\PP_\theta(\mathbf{x}^{[0]} \,|\, \mathbf{x}^{[0,-1]})$ denotes the probability that the pattern at time 0 is $\mathbf{x}^{[0]}$, where $\mathbf{x}^{[0,-1]}$ should be interpreted as an empty history.

We study models where the probability is represented with energy
$E_\theta(\cdot)$ as follows:
\begin{align}
 \PP_\theta(\mathbf{x})
 & = \sum_{\mathbf{\tilde h}} \PP_\theta(\mathbf{x},\mathbf{\tilde h}),
 \intertext{where}
 \PP_\theta(\mathbf{x},\mathbf{h})
 & = \frac{\exp\Big(-E_\theta(\mathbf{x}, \mathbf{h})\Big)}
 {\displaystyle\sum_{\mathbf{\tilde x}} \displaystyle\sum_{\mathbf{\tilde h}}
 \exp\Big(-E_\theta(\mathbf{\tilde x}, \mathbf{\tilde h})\Big)},
\end{align}
the summation with respect to $\mathbf{\tilde x}$ is over all of the
possible binary time-series of length $T$, and the summation with
respect to $\mathbf{\tilde h}$ is over all of the possible hidden
values.

The gradient of $f(\theta)$ can then be represented as follows (see
\eqref{eq:BM:generative:grad}):
\begin{align}
  \nabla f(\theta)
  & = - \E_{\rm target}\left[ \E_\theta\left[ \nabla E_\theta(\boldsymbol{X}, \boldsymbol{H}) \,|\, \boldsymbol{X} \right]\right] 
 + \E_\theta\left[ \nabla E_\theta(\boldsymbol{X}, \boldsymbol{H}) \right],
 \label{eq:time-series:grad}
\end{align}
where $\boldsymbol{X}$ represents the random time-series,
$\boldsymbol{H}$ represents the random hidden values, $\E_\theta$
denotes the expectation with respect to the model distribution
$\PP_\theta$, and $\E_{\rm target}$ denotes the expectation with respect
to the target distribution, which in our case is the empirical
distribution of time-series.  When a single time-series $\mathbf{x}$ is
given as the target, \eqref{eq:time-series:grad} is reduced to
\begin{align}
  \nabla f(\theta)
  & = - \E_\theta\left[ \nabla E_\theta(\mathbf{x}, \boldsymbol{H}) \right] 
 + \E_\theta\left[ \nabla E_\theta(\boldsymbol{X}, \boldsymbol{H}) \right],
\end{align}

In other words, $f(\theta)$ can be maximized by maximizing the sum of
\begin{align}
 f_t(\theta)
 & \equiv \log \PP_\theta(\mathbf{x}^{[t]} \,|\, \mathbf{x}^{[0,t-1]}),
 \intertext{where}
 \PP_\theta(\mathbf{x}^{[t]} \,|\, \mathbf{x}^{[0,t-1]})
& = \sum_{\mathbf{\tilde h}} \PP_\theta(\mathbf{x}^{[t]}, \mathbf{\tilde h} \,|\, \mathbf{x}^{[0,t-1]}) \\
 \PP_\theta(\mathbf{x}^{[t]}, \mathbf{h} \,|\, \mathbf{x}^{[0,t-1]}) 
 & = \frac{\exp\Big(-E_\theta(\mathbf{x}^{[t]},\mathbf{h}\mid\mathbf{x}^{[0,t-1]})\Big)}
 {\displaystyle\sum_{\mathbf{\tilde x}^{[t]}}\sum_{\mathbf{\tilde h}}
 \exp\Big(-E_\theta(\mathbf{\tilde x}^{[t]},\mathbf{\tilde h}\mid\mathbf{x}^{[0,t-1]})\Big)
 },
 \label{eq:time-series:conditional}
\end{align}
and $E_\theta(\mathbf{x}^{[t]},\mathbf{h}\mid\mathbf{x}^{[0,t-1]})$ is
the conditional energy of $(\mathbf{x}^{[t]}, \mathbf{h})$ given
$\mathbf{x}^{[0,t-1]}$.

The gradient of $f_t(\theta)$ is given analogously to $\nabla f(\theta)$:
\begin{align}
  \nabla f_t(\theta)
  & = - \E_{\rm target}\left[ \E_\theta\left[ \nabla E_\theta(\boldsymbol{X}^{[t]}, \boldsymbol{H} \,|\, \boldsymbol{X}^{[t]}, \mathbf{x}^{[0,t-1]}) \right]\right] 
 + \E_\theta\left[ \nabla E_\theta(\boldsymbol{X}^{[t]}, \boldsymbol{H} \mid \mathbf{x}^{[0,t-1]}) \right].
\end{align}
When the target is a single time-series $\mathbf{x}$, we have
\begin{align}
  \nabla f_t(\theta)
 & = - \E_\theta\left[ \nabla E_\theta(\mathbf{x}^{[t]}, \boldsymbol{H} \,|\, \mathbf{x}^{[0,t-1]}) \right]
 + \E_\theta\left[ \nabla E_\theta(\boldsymbol{X}^{[t]}, \boldsymbol{H} \mid \mathbf{x}^{[0,t-1]}) \right].
\end{align}

\section{Non-recurrent Boltzmann machines for time-series}
\label{sec:nonrecurrent}

By \eqref{eq:DyBM:LL}, any model that can represent the conditional probability $\PP_\theta(\mathbf{x}^{[t]} \,|\, \mathbf{x}^{[0,t-1]})$ can be used for time-series.  In this section, we start with a Boltzmann machine that can be used to model a $D$-th order Markov model for an arbitrarily determined $D$.  In $D$-th order Markov models, the conditional probability can be represented as
\begin{align}
\PP_\theta(\mathbf{x}^{[t]} \,|\, \mathbf{x}^{[0,t-1]})
& = \PP_\theta(\mathbf{x}^{[t]} \,|\, \mathbf{x}^{[t-D,t-1]}).
\label{eq:DyBM:Markov}
\end{align}

\subsection{Conditional restricted Boltzmann machines}
\label{sec:CRBM}

\begin{figure}[tb]
\centering
  \begin{minipage}[b]{0.4\linewidth}
  \centering
  \includegraphics[width=\linewidth]{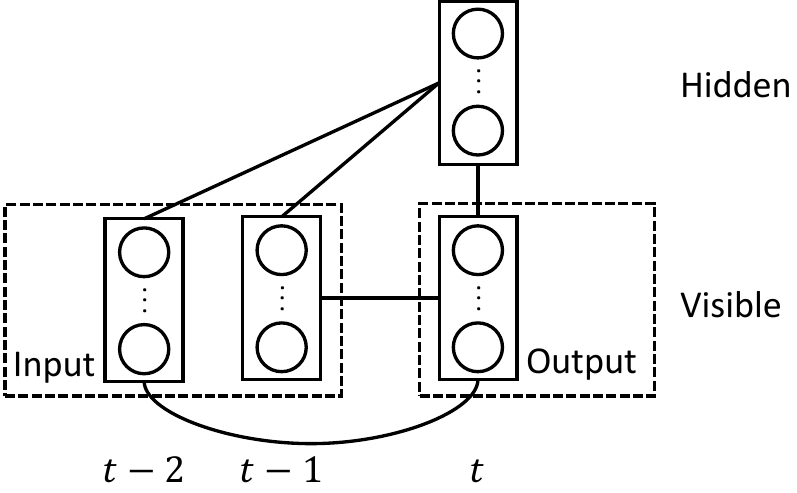}
  \subcaption{Single hidden layer}
  \label{fig:DyBM:CRBM:single}
  \end{minipage}
  \begin{minipage}[b]{0.18\linewidth}
  \ 
  \end{minipage}
  \begin{minipage}[b]{0.4\linewidth}
  \centering
  \includegraphics[width=\linewidth]{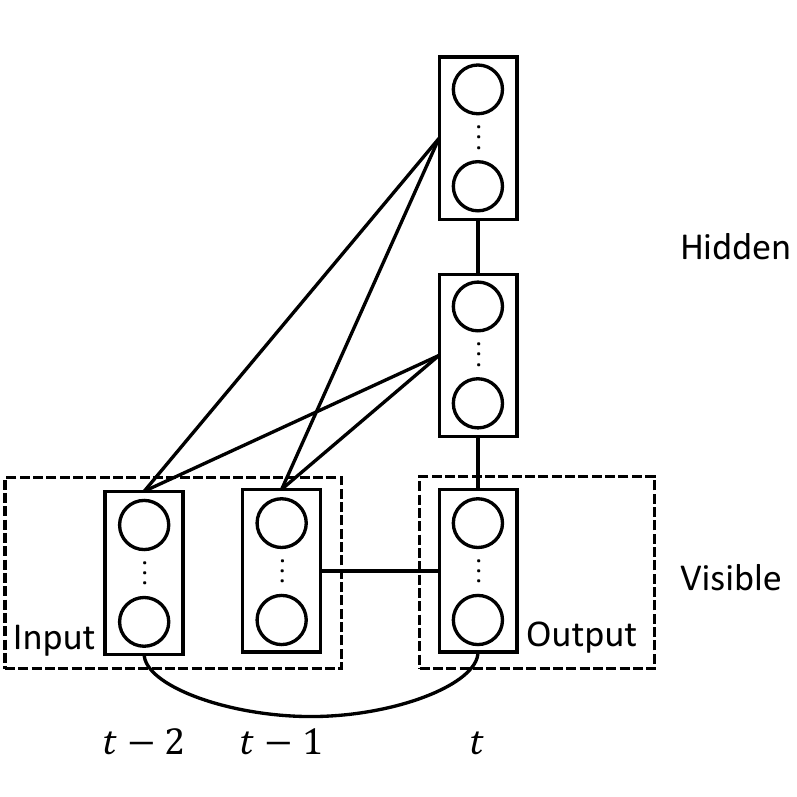}
  \subcaption{Multiple hidden layers}
  \label{fig:DyBM:CRBM:multiple}
  \end{minipage}
\caption{Conditional restricted Boltzmann machines.}
\label{fig:DyBM:CRBM}
\end{figure}

Figure~\ref{fig:DyBM:CRBM:single} shows a particularly structured Boltzmann machine called Conditional Restricted Boltzmann Machine (CRBM) \cite{CRBM}.  A CRBM represents the conditional probability on the right-hand side of \eqref{eq:DyBM:Markov}.  In the figure, we set $D=2$.

The CRBM consists of $D+1$ layers of visible units and a layer of hidden units.  The units within each layer have no connections, but units between different layers may be connected to each other.  Each visible layer corresponds to a pattern at a time $s\in[t-D,t]$.  

The CRBM is a conditional Boltzmann machine shown in Figure~\ref{fig:BM:roles:inout} from \cite{BMsurvey1} but with a particular structure to represent time-series.  The visible layers corresponding to $\mathbf{x}^{[t-D,t-1]}$ are the input, and the visible layer corresponding to $\mathbf{x}^{[t]}$ is the output.  The parameters $\theta$ of the CRBM are independent of $t$.

More formally, the energy of a CRBM is given by
\begin{align}
 E_\theta\big(\mathbf{x}^{[t]}, \mathbf{h} \mid \mathbf{x}^{[t-D,t-1]} \big)
 & = - (\mathbf{b}^{\rm V})^\top \, \mathbf{x}^{[t]}
 - (\mathbf{b}^{\rm H})^\top \, \mathbf{h}
 - \mathbf{h}^\top \, \mathbf{W}^{\rm HV} \, \mathbf{x}^{[t]}
 - \sum_{d=1}^D (\mathbf{x}^{[t-d]})^\top \, \mathbf{W}^{[d]} \, \mathbf{x}^{[t]},
\end{align}
where $\mathbf{x}^{[t]}$ is output, $\mathbf{x}^{[t-D,t-1]}$ is input,
and $\mathbf{h}$ is hidden.

We can then represent the conditional probability as follows (see \eqref{eq:BM:Pyx} from \cite{BMsurvey1}):
\begin{align}
\PP_\theta(\mathbf{x}^{[t]} \,|\, \mathbf{x}^{[t-D,t-1]})
& = \sum_{\mathbf{\tilde h}} 
 \PP_\theta(\mathbf{x}^{[t]},\mathbf{\tilde h} \,|\, \mathbf{x}^{[t-D,t-1]})
 \label{eq:BM:Pyx},
 \intertext{where}
 \PP_\theta(\mathbf{x}^{[t]},\mathbf{\tilde h} \,|\, \mathbf{x}^{[t-D,t-1]})
 & = \frac{\exp\Big(- E_\theta\big(\mathbf{x}^{[t]}, \mathbf{\tilde h} \mid \mathbf{x}^{[t-D,t-1]} \big)\Big)}
 {\displaystyle\sum_{\mathbf{\tilde x}^{[t]}} \exp\Big( - E_\theta\big(\mathbf{\tilde x}^{[t]}, \mathbf{\tilde h} \mid \mathbf{x}^{[t-D,t-1]} \big) \Big)},
\end{align}
and the summation with respect to $\mathbf{\tilde h}$ is over all of the
possible binary hidden patterns, and the summation with respect to
$\mathbf{\tilde x}^{[t]}$ is defined analogously.

One can then learn the parameters $\theta=(\mathbf{b}^{\rm V},
\mathbf{h}^{\rm h}, \mathbf{W}^{\rm HV}, \mathbf{W}^{[1]}, \ldots,
\mathbf{W}^{D})$ of the model by following a gradient-based method in
Section~\ref{sec:BM:discriminative} from \cite{BMsurvey1}.

\subsection{Extensions of conditional restricted Boltzmann machines}

The CRBM has been extended in various ways.  Taylor et al.\ study a CRBM with multiple layers of hidden units \cite{CRBM} (see Figure~\ref{fig:DyBM:CRBM:multiple}).  Memisevic and Hinton study a CRBM extended with three-way interactions ({\it i.e.}, a higher order Boltzmann machine), which they refer to as a gated CRBM \cite{GatedCRBM}.  Specifically, the energy of the gated CRBM involves
\begin{align}
 - \sum_{i,j,k} w_{i,j,k} \, x_i \, y_j \, h_k,
\end{align}
where $\mathbf{x}$ denotes input values, $\mathbf{y}$ denotes output values, and $\mathbf{h}$ denotes hidden values.  A drawback of the gated CRBM is its increased number of parameters due to the three-way interactions.  Taylor and Hinton study a factored CRBM, where the three-way interaction is represented with a reduced number of parameters as follows \cite{TayHin09}:
\begin{align}
 - \sum_f \sum_{i,j,k} w_{i,f}^\mathbf{v} \, w_{j,f}^\mathbf{y} \, w_{k,f}^\mathbf{h} \, x_i \, y_j \, h_k,
\end{align}
where the summation with respect to $f$ is over a set of factors under consideration.

\section{Boltzmann machines for time-series with recurrent structures}
\label{sec:recurrent}

\subsection{Spiking Boltzmann machines}

\begin{figure}[t]
 \begin{minipage}[b]{0.5\linewidth}
  \centering
  \includegraphics[width=\linewidth]{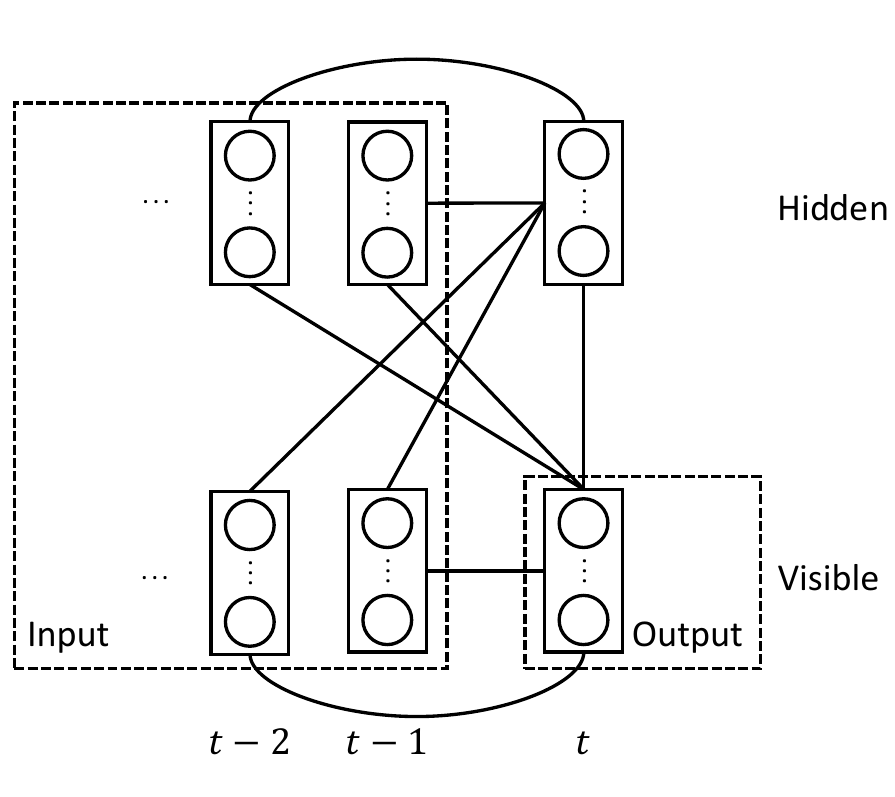}
  (a) Structure of a spiking Boltzmann machine
\end{minipage}
\begin{minipage}[b]{0.5\linewidth}
\centering
\includegraphics[width=\linewidth]{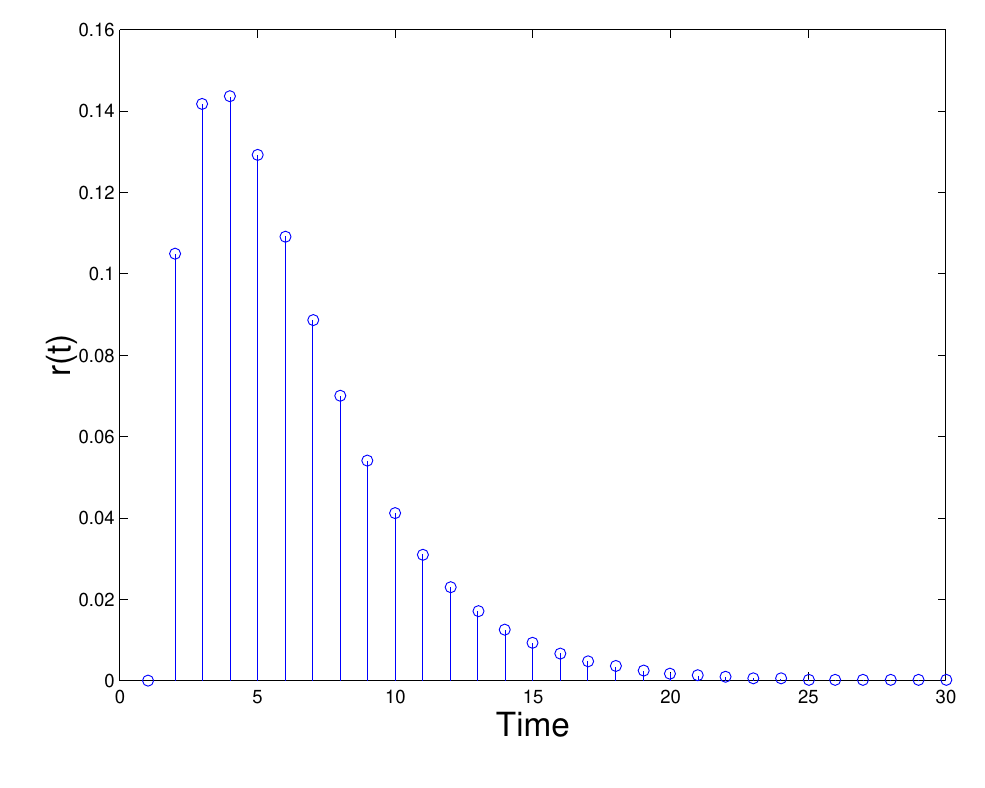}
(b) $r(\cdot)$ used in \cite{HinBro99}
\end{minipage}
\caption{A spiking Boltzmann machine studied in \cite{HinBro99}.  In (b), 
we use the figure in the version available at 
http://www.cs.toronto.edu/$\sim$fritz/absps/nips00-ab.pdf.}
\label{fig:DyBM:spiking}
\end{figure}

A spiking Boltzmann machine studied in \cite{HinBro99} can be shown to
be essentially equivalent to the Boltzmann machine illustrated in
Figure~\ref{fig:DyBM:spiking}.  This Boltzmann machine consists of
input units, output units, and hidden units.  The input units represent
historical values of visible units and hidden units.  Although hidden
units are random and cannot be simply given as input, Hinton and Brown
make the approximation of using sampled values
$\boldsymbol{H}^{(-\infty,t-1]}(\omega)$ as the input hidden units
\cite{HinBro99}.

Specifically, given the visible values $\mathbf{x}^{[<t]} \equiv \mathbf{x}^{(-\infty,t-1]}$ and
sampled hidden values $\boldsymbol{H}^{[<t]}(\omega)$ up to
time $t-1$, the energy with the visible values $\mathbf{x}^{[t]}$ and
hidden values $\mathbf{h}^{[t]}$ at time $t$ can be represented as
follows:
\begin{align}
E_\theta(\mathbf{x}^{[t]},\mathbf{h}^{[t]} \mid \mathbf{x}^{[<t]}, \boldsymbol{H}^{[<t]}(\omega))
& = 
- (\mathbf{b}^{\rm V})^\top \, \mathbf{x}^{[t]}
- (\mathbf{b}^{\rm H})^\top \, \mathbf{h}^{[t]}
- (\mathbf{h}^{[t]})^\top \, r(\tau) \, \mathbf{W}^{\rm HV} \, \mathbf{x}^{[t]} \notag\\
 & \hspace{4mm}
 - \sum_{\tau=1}^\infty (\boldsymbol{H}^{[t-\tau]}(\omega))^\top \, r(\tau) \, \mathbf{W}^{\rm HH} \, \mathbf{h}^{[t]}
- \sum_{\tau=1}^\infty (\mathbf{x}^{[t-\tau]})^\top \, r(\tau) \, \mathbf{W}^{\rm VH} \, \mathbf{h}^{[t]} \notag\\
& \hspace{4mm}
- \sum_{\tau=1}^\infty (\boldsymbol{H}^{[t-\tau]}(\omega))^\top \, r(\tau) \, \mathbf{W}^{\rm HV} \, \mathbf{x}^{[t]}
- \sum_{\tau=1}^\infty (\mathbf{x}^{[t-\tau]})^\top \, r(\tau) \, \mathbf{W}^{\rm VV} \, \mathbf{x}^{[t]},
 \label{eq:DyBM:spiking:energy}
\end{align}
where $r(\cdot)$ is an arbitrarily chosen function and is not the target 
of learning.  Namely, the Boltzmann machine has an infinite number of 
units but can be characterized by a finite number of parameters 
$\theta\equiv(\mathbf{b}^{\rm V},\mathbf{b}^{\rm H},\mathbf{W}^{\rm 
VV},\mathbf{W}^{\rm VH},\mathbf{W}^{\rm HV},\mathbf{W}^{\rm HH})$. 
Figure~\ref{fig:DyBM:spiking}(b) shows the specific $r(\cdot)$ used in 
\cite{HinBro99}\footnote{Although it 
is not clear from the descriptions in \cite{HinBro99}, the labels in the 
horizontal axis should probably be shifted by one, so that $r(0)=0$, 
$r(1)\approx 0.1$, and so on.}.

Notice that the Boltzmann machine in Figure~\ref{fig:DyBM:spiking} can 
be seen as a restricted Boltzmann machine (RBM) whose bias and weight can depend 
on $\mathbf{x}^{[<t]}$ and $\boldsymbol{H}^{[<t]}(\omega)$, 
because \eqref{eq:DyBM:spiking:energy} can be represented as 
\begin{align}
E_\theta(\mathbf{x}^{[t]},\mathbf{h}^{[t]} \mid \mathbf{x}^{[<t]}, \mathbf{H}^{[<t]}(\omega))
& = 
- \mathbf{b}^{\rm H}(t,\omega) \, \mathbf{h}^{[t]}
- \mathbf{b}^{\rm V}(t,\omega) \, \mathbf{x}^{[t]}
- \mathbf{h}^{[t]} \, \mathbf{W} \, \mathbf{x}^{[t]},
\end{align}
where $\mathbf{b}^{\rm H}(t,\omega)$ is the time-varying bias for hidden units, 
$\mathbf{b}^{\rm V}(t,\omega)$ is the time-varying bias for visible units, 
and $\mathbf{W}$ is the weight between visible units and hidden units:
\begin{align}
 \mathbf{b}^{\rm H}(t,\omega)
 & \equiv \mathbf{b}^{\rm H} 
	+ \sum_{\tau=1}^\infty (\boldsymbol{H}^{[t-\tau]}(\omega))^\top \, r(\tau) \, \mathbf{W}^{\rm HH} 
	+ \sum_{\tau=1}^\infty (\mathbf{x}^{[t-\tau]})^\top \, r(\tau) \, \mathbf{W}^{\rm VH} \\
\mathbf{b}^{\rm V}(t,\omega) & \equiv
	\mathbf{b}^{\rm V}
	+ \sum_{\tau=1}^\infty (\boldsymbol{H}^{[t-\tau]}(\omega))^\top \, r(\tau) \, \mathbf{W}^{\rm HV}
	+ \sum_{\tau=1}^\infty (\mathbf{x}^{[t-\tau]})^\top \, r(\tau) \, \mathbf{W}^{\rm VV} \\
\mathbf{W} & \equiv
	r(0) \, \mathbf{W}^{\rm HV}.
\end{align}

We can then represent the conditional probability as follows:
\begin{align}
\PP_\theta(\mathbf{x}^{[t]} \,|\, \mathbf{x}^{[<t]}, \boldsymbol{H}^{(-\infty, t-1}(\omega))
& = \sum_{\mathbf{\tilde h}^{[t]}} 
 \PP_\theta\big(\mathbf{x}^{[t]},\mathbf{\tilde h}^{[t]} \,|\, \mathbf{x}^{[<t]}, \boldsymbol{H}^{(-\infty, t-1]}(\omega)\big)
 \intertext{where}
 \PP_\theta(\mathbf{x}^{[t]},\mathbf{\tilde h}^{[t]} \,|\, \mathbf{x}^{[<t]}, \boldsymbol{H}^{[<t]}(\omega))
 & = \frac{\exp\Big(- E_\theta\big(\mathbf{x}^{[t]}, \mathbf{\tilde h}^{[t]} \mid \mathbf{x}^{[<t]}, \boldsymbol{H}^{(-\infty, t-1]}(\omega) \big)\Big)}
 {\displaystyle\sum_{\mathbf{\tilde x}^{[t]}} \exp\Big(
 - E_\theta\big(\mathbf{\tilde x}^{[t]}, \mathbf{\tilde h}^{[t]}
 \mid \mathbf{x}^{[<t]}, \boldsymbol{H}^{[<t]}(\omega) \big)
 \Big)}.
\end{align}

We now discuss the choice of $r(0)=0$, which appears to be the case in 
Figure~\ref{fig:DyBM:spiking}(b).  In this case, the energy is reduced to
\begin{align}
E_\theta(\mathbf{x}^{[t]},\mathbf{h}^{[t]} \mid \mathbf{x}^{[<t]}, \boldsymbol{H}^{[<t]}(\omega))
& = 
- (\mathbf{b}^{\rm H}(t,\omega))^\top \, \mathbf{h}^{[t]}
- (\mathbf{b}^{\rm V}(t,\omega))^\top \, \mathbf{x}^{[t]}.
\end{align}
Because there are no connections between visible units and hidden units at time $t$, 
the hidden values at $t$ do not affect the distribution of the visible values at $t$.  
The only role of the hidden units is that the sampled hidden values are used to update the time-varying bias, $\mathbf{b}^{\rm V}(s,\omega)$ 
and $\mathbf{b}^{\rm H}(s,\omega)$ for $s>t$.  A problem is that there is no mechanism 
that allows us to learn appropriate values of $\mathbf{W}^{\rm VH}$ and 
$\mathbf{W}^{\rm HH}$ until we observe succeeding visible values.  Namely, the hidden values $\mathbf{h}^{[t]}$ are 
sampled with the dependency on $\mathbf{W}^{\rm VH}$ and 
$\mathbf{W}^{\rm HH}$, but whether the sampled hidden values are good or 
not can only be known when those hidden values are used as input.  This 
helps us to learn appropriate values of $\mathbf{W}^{\rm HV}$, but not 
$\mathbf{W}^{\rm VH}$ or $\mathbf{W}^{\rm HH}$.  See 
\cite{BidirectionalDyBM} for further discussion.

\subsection{Temporal restricted Boltzmann machines}

\begin{figure}[tb]
\centering
  \begin{minipage}[b]{0.4\linewidth}
  \centering
  \includegraphics[width=\linewidth]{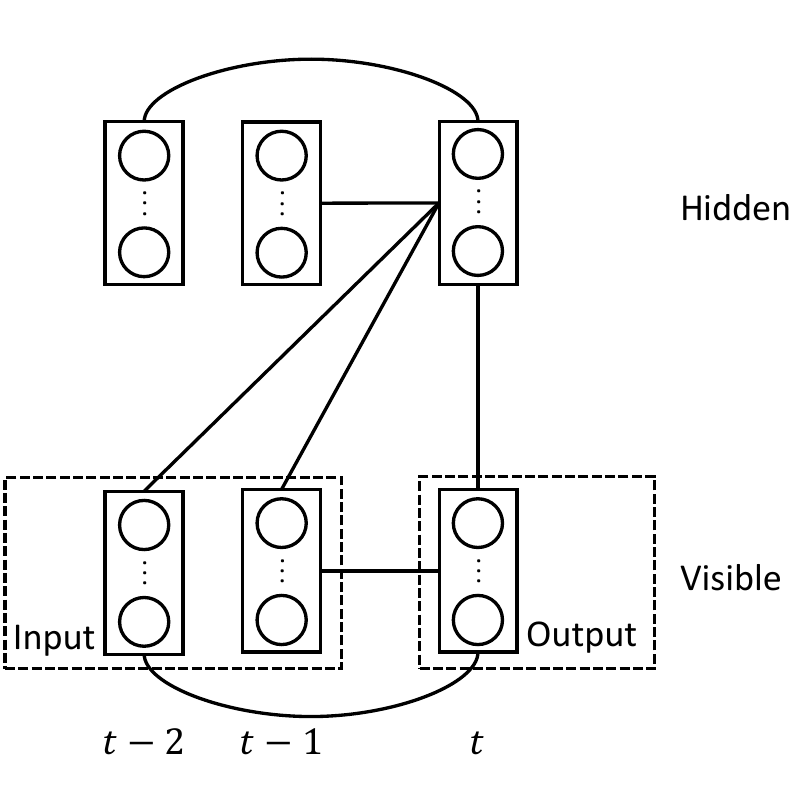}
  \subcaption{TRBM}
  \label{fig:DyBM:TRBM:single}
  \end{minipage}
  \begin{minipage}[b]{0.18\linewidth}
  \ 
  \end{minipage}
  \begin{minipage}[b]{0.4\linewidth}
  \centering
  \includegraphics[width=\linewidth]{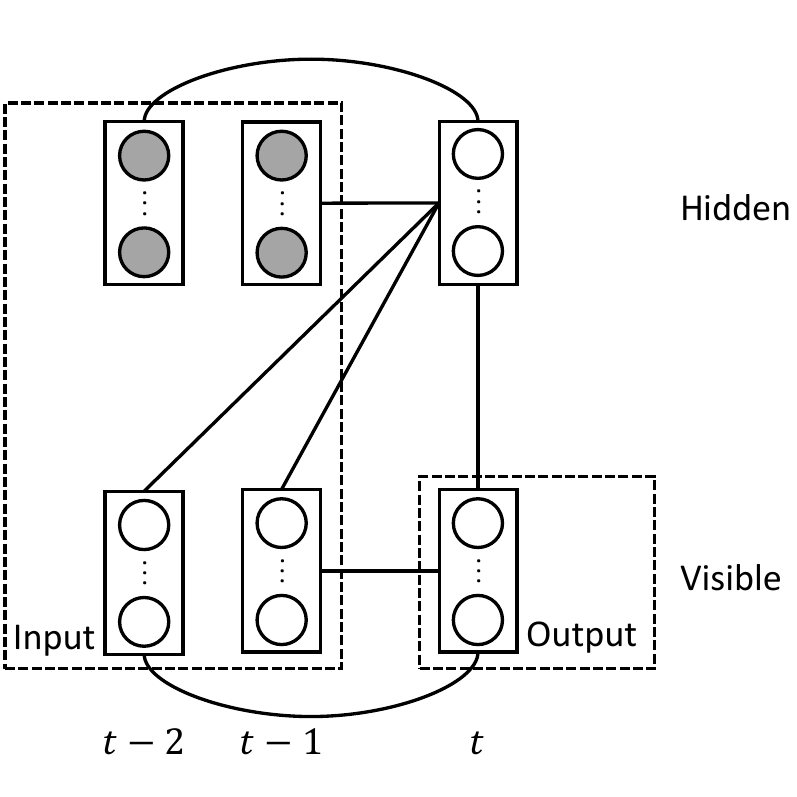}
  \subcaption{TRBM with approximations in \cite{TRBM}}
  \label{fig:DyBM:TRBM-approx}
  \end{minipage}
\caption{Temporal restricted Boltzmann machines.  In (b), the gray circles indicate that expected values are used for the input hidden units.}
\label{fig:DyBM:TRBM}
\end{figure}

Sutskever and Hinton study a model related to a CRBM, which they refer to as a temporal restricted Boltzmann machine (TRBM) \cite{TRBM}.  While a CRBM defines the conditional distribution of the (visible and hidden) values at time $t$ given only the visible values from time $t-D$ to $t-1$, a TRBM defines the corresponding conditional probability given both the visible values and the hidden values from time $t-D$ to $t-1$.  See Figure~\ref{fig:DyBM:TRBM:single}.  Similar to the CRBM, the parameters $\theta$ of the TRBM do not depend on time $t$.

Unlike the CRBM, the TRBM is not a conditional RBM.  This is because the TRBM with a single parameter is used for every $t$, and the distribution of hidden values is shared among those TRBM at varying $t$.  In particular, hidden values of the TRBM can depend on the future visible values.  

Because this dependency makes learning and inference hard, it is ignored in \cite{TRBM}.  Namely, the values at each time $t$ is conditionally independent of the values after time $t$ given the values at and before time $t$.  In particular, the distribution of the hidden values at time $t$ is completely determined by the visible values up to time $t$.  The distribution of the hidden values before time $t$ can thus be considered as input when we use the TRBM to define the conditional distribution of the values at time $t$ (see Figure~\ref{fig:DyBM:TRBM-approx}).  For each sampled values of hidden units, TRBM in \ref{fig:DyBM:TRBM-approx} is a CRBM.  

Furthermore, in \cite{TRBM}, the expected values (see Section~\ref{sec:BM:real:meanfield} from \cite{BMsurvey1}) are used for the hidden values.  Then the input hidden units in Figure~\ref{fig:DyBM:TRBM-approx} takes real values in $[0,1]$ that are completely determined by the visible values before time $t$.

More formally, with the approximations in \cite{TRBM}, the TRBM with parameter $\theta$ defines the probability distribution over the time-series of visible and hidden values as follows:
\begin{align}
\PP_\theta(\mathbf{x})
 & = \prod_{t=0}^T \sum_{\mathbf{\tilde h}^{[t]}}
 \PP_\theta(\mathbf{x}^{[t]},\mathbf{\tilde h}^{[t]} \,|\, \mathbf{x}^{[t-D,t-1]},\mathbf{r}^{[t-D,t-1]}),
\intertext{where}
 \PP_\theta(\mathbf{x}^{[t]},\mathbf{\tilde h}^{[t]} \,|\, \mathbf{x}^{[t-D,t-1]},\mathbf{r}^{[t-D,t-1]})
 & = \frac{\exp\Big(-E_\theta(\mathbf{x}^{[t]},\mathbf{\tilde h}^{[t]} \,|\, \mathbf{x}^{[t-D,t-1]},\mathbf{r}^{[t-D,t-1]})\Big)}
 {\displaystyle\sum_{\mathbf{\tilde x}^{[t]}} \exp\Big(-E_\theta(\mathbf{\tilde x}^{[t]},\mathbf{\tilde h}^{[t]} \,|\, \mathbf{x}^{[t-D,t-1]},\mathbf{r}^{[t-D,t-1]})\Big)}
 \label{eq:TRBM}
\end{align}
is the conditional
distribution defined by the Boltzmann machine shown in
Figure~\ref{fig:DyBM:TRBM-approx}, where $\mathbf{r}^{[t-D,t-1]}$ are
expected hidden values.  Specifically,  \eqref{eq:TRBM} is used to compute
\begin{align}
 \mathbf{r}^{[t]}
 & = \E_\theta[ \boldsymbol{H}^{[t]} \mid \mathbf{x}^{[0,t]} ],
 \label{eq:TRBM:E}
\end{align}
which is subsequently used with \eqref{eq:TRBM} for $t\leftarrow t+1$, where the expectation in \eqref{eq:TRBM:E} is with respect to
\begin{align}
 \PP_\theta( \mathbf{h}^{[t]} \mid \mathbf{x}^{[0,t-1]})
 & = \frac{ \PP_\theta( \mathbf{x}^{[t]}, \mathbf{h}^{[t]} \mid \mathbf{x}^{[t-D,t-1]}, \mathbf{r}^{[t-D,t-1]}) }
 { \displaystyle\sum_{\mathbf{\tilde x}^{[t]}} \PP_\theta( \mathbf{\tilde x}^{[t]}, \mathbf{h}^{[t]} \mid \mathbf{x}^{[t-D,t-1]}, \mathbf{r}^{[t-D,t-1]}) }.
\end{align}

Notice that $\mathbf{r}^{[t]}$ can be computed from $\mathbf{x}^{[0,t]}$
in a deterministic manner with dependency on $\theta$.  However, this
dependency on $\theta$ is ignored in learning TRBMs.


\subsection{Recurrent temporal restricted Boltzmann machines}

\begin{figure}[tb]
\centering
  \includegraphics[width=0.4\linewidth]{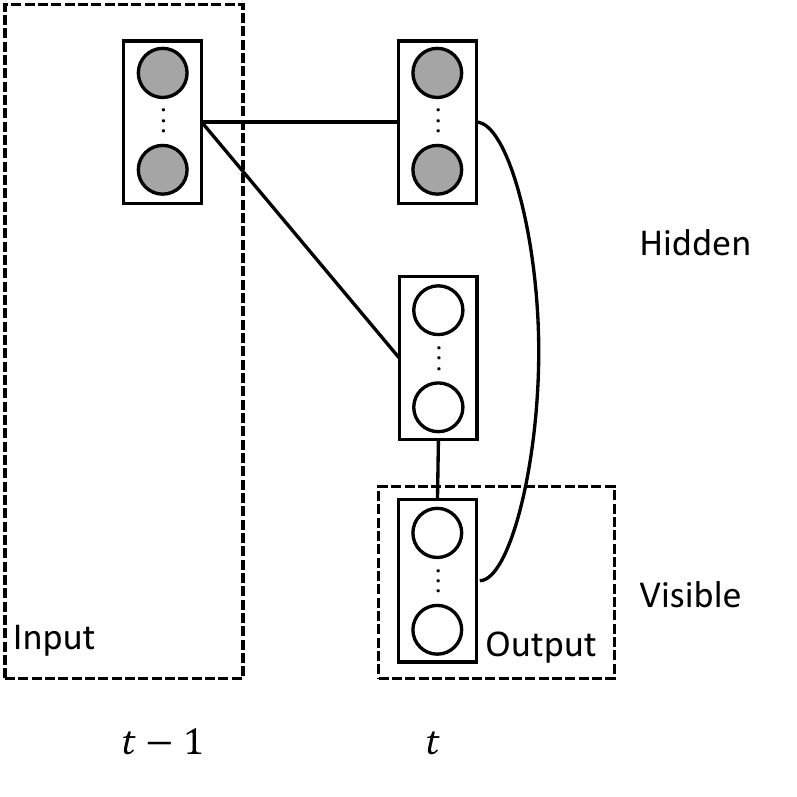}
\caption{A recurrent temporal restricted Boltzmann machine.}
\label{fig:DyBM:RTRBM}
\end{figure}

To overcome the intractability of the TRBM without approximations, Sutskever et al.\ study a refined model of TRBM, which they refer to as a recurrent temporal restricted Boltzmann machine (RTRBM) \cite{RTRBM}.  The RTRBM simplifies the TRBM by removing connections between visible layers and connections between hidden layers that are separated by more than one lag.  This means that the (visible and hidden) values at time $t$ are conditionally independent of the the visible values before time $t$ and the hidden values before time $t-1$ given the hidden values at time $t-1$.  Similar to the approximation made for the TRBM in Figure~\ref{fig:DyBM:TRBM-approx}, the RTRBM uses the expected values for the hidden values at time $t-1$ but defines the conditional distribution of the (visible and hidden) values at time $t$ over their binary values.  See Figure~\ref{fig:DyBM:RTRBM}.

More formally, let $\mathbf{r}^{[t-1]}$ denote the expected values of the hidden units at time $t-1$:
\begin{align}
\mathbf{r}^{[t-1]}
\equiv
\E_\theta\big[ \mathbf{H}^{[t-1]} \,|\, \mathbf{x}^{[0,t-1]} \big],
\end{align}
where $\mathbf{H}^{[t-1]}$ is the random vector representing the hidden values at time $t$, and $\E_\theta[\cdot \,|\, \mathbf{x}^{[0,t-1]} ]$ represents the conditional expectation given the visible values up to time $t-1$.  The probability distribution of the values at time $t$ is then given by
\begin{align}
\PP_\theta(\mathbf{x}^{[t]},\mathbf{h}^{[t]} \,|\, \mathbf{r}^{[t-1]})
& = \frac{\exp\Big(-E_\theta(\mathbf{x}^{[t]},\mathbf{h}^{[t]}\,|\,\,\mathbf{r}^{[t-1]})\Big)}
	{\displaystyle\sum_{\mathbf{\tilde x},\mathbf{\tilde h}} \exp\Big(-E_\theta(\mathbf{\tilde x}^{[t]},\mathbf{\tilde h}^{[t]}\,|\,\,\mathbf{r}^{[t-1]})\Big)},
\end{align}
where the conditional energy is given by
\begin{align}
E_\theta(\mathbf{x}^{[t]},\mathbf{h}^{[t]}\,|\,\,\mathbf{r}^{[t-1]})
\equiv
- (\mathbf{b}^{\rm V})^\top \mathbf{x}^{[t]}
- (\mathbf{b}^{\rm H})^\top \mathbf{h}^{[t]}
- (\mathbf{r}^{[t-1]})^\top \mathbf{U} \, \mathbf{h}^{[t]}
- (\mathbf{x}^{[t]})^\top \mathbf{W} \, \mathbf{h}^{[t]}
\end{align}
for parameters
\begin{align}
 \theta \equiv (\mathbf{b}^{\rm V}, \mathbf{b}^{\rm H}, \mathbf{W}, \mathbf{U}),
\end{align}
where $\mathbf{b}^{\rm V}$ is the bias for visible units, $\mathbf{b}^{\rm H}$ is the bias for hidden units,
$\mathbf{W}$ is the weight matrix between visible units and hidden units, and
$\mathbf{U}$ is the weight matrix between previous expected value of hidden units (at $t-1$) and hidden units (at $t$).

\subsubsection{Inference}

The marginal conditional distribution of visible values at time $t-1$ can then be represented as follows:
\begin{align}
\PP_\theta(\mathbf{x}^{[t]} \,|\, \mathbf{r}^{[t-1]})
& = \frac{\exp\Big(-F_\theta(\mathbf{x}^{[t]}\,|\,\,\mathbf{r}^{[t-1]})\Big)}
	{\displaystyle\sum_{\mathbf{\tilde x}^{[t]}} \exp\Big(-F_\theta(\mathbf{\tilde x}^{[t]}\,|\,\,\mathbf{r}^{[t-1]})\Big)},
\label{eq:DyBM:RTRBM:P}
\end{align}
where the conditional free-energy is given by
\begin{align}
F_\theta(\mathbf{x}^{[t]}\,|\,\,\mathbf{r}^{[t-1]})
\equiv
- \log \sum_{\mathbf{\tilde h^{[t]}}}
\exp\Big( 
	-E_\theta(\mathbf{x}^{[t]},\mathbf{\tilde h}^{[t]} 
		\,|\, \mathbf{r}^{[t-1]}) 
\Big).
\label{eq:DyBM:RTRBM:F}
\end{align}

Once the visible values at time $t$ is given, the conditional probability distribution of the hidden values at time $t$ can be represented as follows:
\begin{align}
\PP_\theta(\mathbf{h}^{[t]} \,|\, \mathbf{r}^{[t-1]},\mathbf{x}^{[t]})
& = \frac{\exp\Big(-E_\theta(\mathbf{h}^{[t]}\,|\,\,\mathbf{r}^{[t-1]},\mathbf{x}^{[t]})\Big)}
	{\displaystyle\sum_{\mathbf{\tilde h}^{[t]}} \exp\Big(-E_\theta(\mathbf{\tilde h}^{[t]}\,|\,\,\mathbf{r}^{[t-1]},\mathbf{x}^{[t]})\Big)},
\end{align}
where the $(\mathbf{b}^{\rm V})^\top \mathbf{x}^{[t]}$ term is canceled out between the numerator and the denominator, and the conditional energy is given by
\begin{align}
E_\theta(\mathbf{h}^{[t]}\,|\,\,\mathbf{r}^{[t-1]},\mathbf{x}^{[t]})
\equiv
- \Big(\mathbf{b}^{\rm H} 
	+ \mathbf{U}^\top \, \mathbf{r}^{[t-1]}
	+ \mathbf{W}^\top \, \mathbf{x}^{[t]}
\Big)^\top \mathbf{h}^{[t]}.
\end{align}
By Corollary~\ref{corollary:BM:independence} from \cite{BMsurvey1}, the hidden values at time $t$ are conditionally independent of each other given $\mathbf{r}^{[t-1]}$ and $\mathbf{x}^{[t]}$:
\begin{align}
\PP_\theta(\mathbf{h}^{[t]} \,|\, \mathbf{r}^{[t-1]},\mathbf{x}^{[t]})
& = \prod_i \PP_\theta(h_i^{[t]} \,|\, \mathbf{r}^{[t-1]},\mathbf{x}^{[t]}), \\
\intertext{where}
\PP_\theta(h_i^{[t]} \,|\, \mathbf{r}^{[t-1]},\mathbf{x}^{[t]})
& = 
\frac{\exp\big( -b_i^{[t]} \, h_i^{[t]} \big)}
	{1 + \exp\big( -b_i^{[t]} \big)},
\end{align}
where $b_i^{[t]}$ is the $i$-th element of 
\begin{align}
\mathbf{b}^{[t]}
& \equiv
\mathbf{b}^{\rm H} 
	+ \mathbf{U}^\top \, \mathbf{r}^{[t-1]}
	+ \mathbf{W}^\top \, \mathbf{x}^{[t]} \label{eq:DyBM:RTRBM:b} \\
\intertext{for $t \ge 1$, and}
\mathbf{b}^{[0]}
& \equiv
\mathbf{b}^{\rm init} 
	+ \mathbf{W}^\top \, \mathbf{x}^{[0]}.
\label{eq:DyBM:RTRBM:b0}
\end{align}
where we now follow \cite{RTRBM} and allow the hidden units at time $0$ to have own bias $\mathbf{b}^{\rm init}$ that can differ from $\mathbf{b}^{\rm H}$.  The expected values are thus given by 
\begin{align}
\mathbf{r}^{[t]} 
& = 
\frac{1}{1 + \exp\big( \mathbf{b}^{[t]} \big)},
\label{eq:DyBM:RTRBM:barh}
\end{align}
where the operations are defined elementwise.

\subsubsection{Learning}

\begin{figure}[tb]
\centering
  \includegraphics[width=0.8\linewidth]{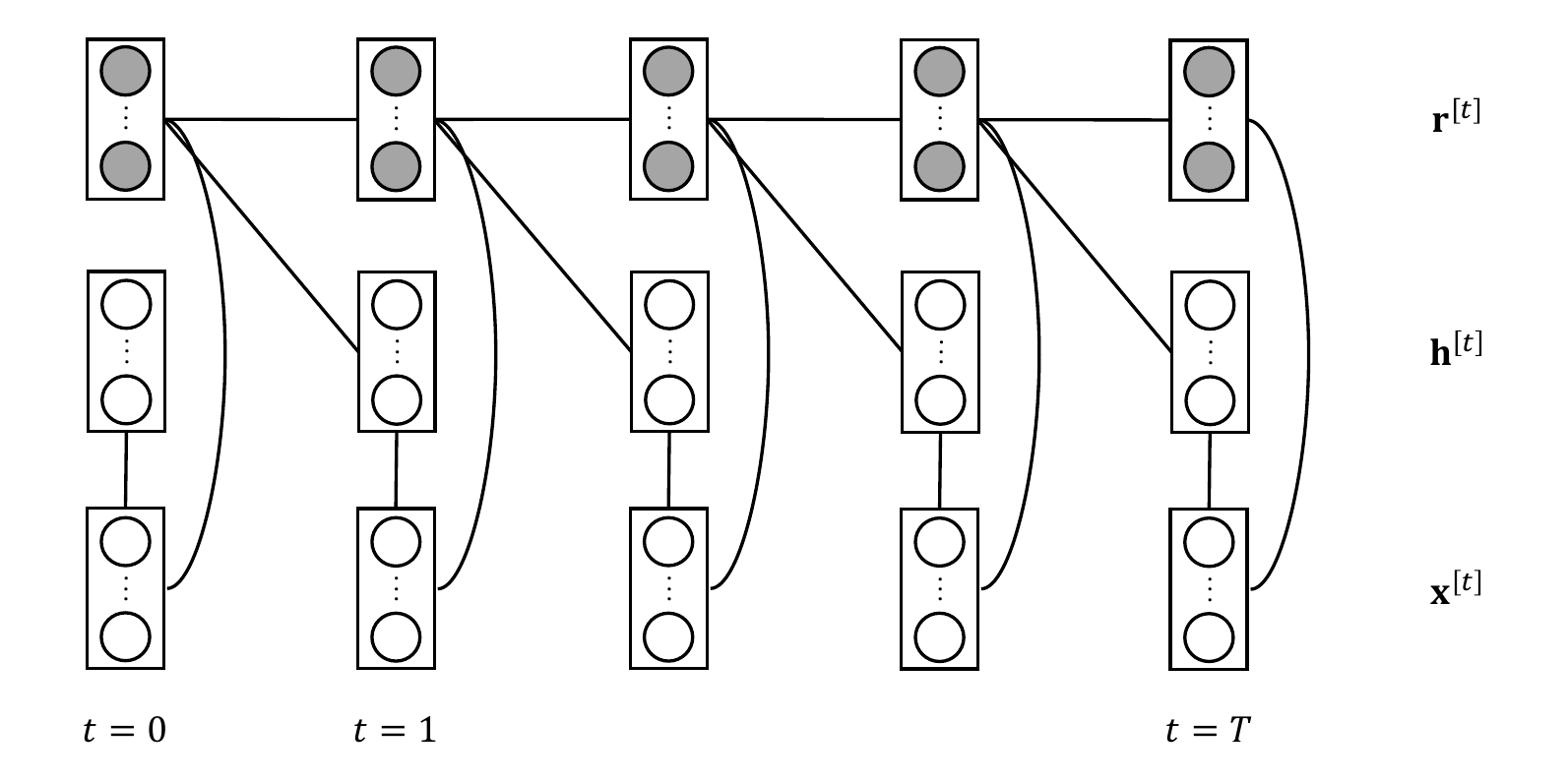}
\caption{A recurrent temporal restricted Boltzmann machine unfolded through time, where $T=4$.}
\label{fig:DyBM:RTRBM-unfold}
\end{figure}

The parameters of an RTRBM can be trained through back propagation through time, analogous to recurrent neural networks, but with contrastive divergence.  To understand how we can train RTRBMs, Figure~\ref{fig:DyBM:RTRBM-unfold} shows an RTRBM unfolded through time.  Recall that the expected values of hidden units are deterministically updated from $\mathbf{r}^{[t-1]}$ to $\mathbf{r}^{[t]}$ according to \eqref{eq:DyBM:RTRBM:b}-\eqref{eq:DyBM:RTRBM:barh}.  Hence, $\mathbf{r}^{[t]}$ can be understood as hidden values of a recurrent neural network (RNN) \cite{RNN}.  An RTRBM can then be seen as an RNN but gives an RBM as an output instead of real values, which would be given as an output from the standard RNN.  We will derive the learning rule of the RTRBM, closely following \cite{SRTRBM} but using our notations.

When an RTRBM is unfolded through time, its energy can be represented as follows:
\begin{align}
E_\theta(\mathbf{x},\mathbf{h})
& =
- \sum_{t=0}^T (\mathbf{b}^{\rm V})^\top \mathbf{x}^{[t]}
- (\mathbf{b}^{\rm init})^\top \mathbf{h}^{[0]}
- \sum_{t=1}^T (\mathbf{b}^{\rm H})^\top \mathbf{h}^{[t]}
- \sum_{t=0}^T (\mathbf{x}^{[t]})^\top \mathbf{W} \, \mathbf{h}^{[t]}
- \sum_{t=1}^T (\mathbf{r}^{[t-1]})^\top \mathbf{U} \, \mathbf{h}^{[t]}.
\label{eq:DyBM:RTRBM-energy}
\end{align}

By \eqref{eq:time-series:grad}, we can maximize the log-likelihood of a given time-series $\mathbf{x}$ with a gradient-based approach.  What we need in \eqref{eq:time-series:grad} is the gradient of the energy with respect to the parameter.  A caveat is that the energy in \eqref{eq:DyBM:RTRBM-energy} depends on $\mathbf{r}^{[\cdot]}$, which in turn depends on $\theta$ in a recursive manner.  Also, expectation with respect to $\PP_theta$ in \eqref{eq:time-series:grad} needs to be computed with approximation such as contrastive divergence (see Section~\ref{sec:BM:CD}).

We first study the last term of \eqref{eq:DyBM:RTRBM-energy}.  Let
\begin{align}
Q_s 
& \equiv \sum_{t=s}^T (\mathbf{r}^{[t-1]})^\top \mathbf{U} \, \mathbf{h}^{[t]}\\
& = (\mathbf{r}^{[s-1]})^\top \mathbf{U} \, \mathbf{h}^{[s]} + Q_{s+1},
\end{align}
for $s\in[1,T]$, where $Q_{T+1}\equiv 0$, so that $Q \equiv Q_1$ is the last term of \eqref{eq:DyBM:RTRBM-energy}.  Taking the partial derivative with respect to $r_i^{[s-1]}$, we obtain
\begin{align}
\frac{\partial Q_s}{\partial r_i^{[s-1]}} 
& = 
\frac{\partial}{\partial r_i^{[s-1]}} (\mathbf{r}^{[s-1]})^\top \mathbf{U} \, \mathbf{h}^{[s]} 
+ \sum_j \frac{\partial r_j^{[s]}}{\partial r_i^{[s-1]}} \frac{\partial Q_{s+1}}{\partial r_j^{[s]}} \\
& = 
\mathbf{U}_{i,:} \, \mathbf{h}^{[s]} 
+ \sum_j r_j^{[s]} (1-r_j^{[s]}) \, u_{i,j} \, \frac{\partial Q_{s+1}}{\partial r_j^{[s]}},
\end{align}
where $\mathbf{U}_{i,:}$ denotes the $i$-th row of $\mathbf{U}$, $u_{i,j}$ denotes the $(j,i)$-th element of $\mathbf{U}$, and the last equality follows from \eqref{eq:DyBM:RTRBM:b}-\eqref{eq:DyBM:RTRBM:barh}.  In vector-matrix notations, we can write
\begin{align}
\nabla_{\mathbf{r}^{[s-1]}} Q_s
& = 
\mathbf{U} \, \big(
	\mathbf{h}^{[s]} 
	+
	\mathbf{r}^{[s]} \cdot (1-\mathbf{r}^{[s]}) \cdot \nabla_{\mathbf{r}^{[s]}} Q_{s+1}
	\big),
\end{align}
where $\cdot$ denotes elementwise multiplication.  Because $Q_s$ is not a function of $\mathbf{r}^{[0]}, \ldots, \mathbf{r}^{[s-2]}$, we have 
\begin{align}
\nabla_{\mathbf{r}^{[s-1]}} Q = \nabla_{\mathbf{r}^{[s-1]}} Q_s.
\end{align}
Therefore, the partial derivative of $Q$ with respect to $\mathbf{r}^{[s-1]}$ is given recursively as follows:
\begin{align}
\nabla_{\mathbf{r}^{[s-1]}} Q
& = 
\mathbf{U} \, \big(
	\mathbf{h}^{[s]} 
	+
	\mathbf{r}^{[s]} \cdot (1-\mathbf{r}^{[s]}) \cdot \nabla_{\mathbf{r}^{[s]}} Q
	\big)
\label{eq:DyBM:RTRBM-Q}
\intertext{for $s=1,\ldots,T$ and}
\nabla_{\mathbf{r}^{[T]}} Q & = \mathbf{0}.
\label{eq:DyBM:RTRBM-QT}
\end{align}

We now take the derivative of $Q$ with respect to the parameters in $\theta$, starting with $\mathbf{U}$:
\begin{align}
\frac{{\rm d} Q}{{\rm d} u_{i,j}}
& = \sum_{t=0}^T \sum_k \frac{\partial r_k^{[t]}}{\partial u_{i,j}} \, \frac{\partial Q}{\partial r_k^{[t]}}
+ \frac{\partial Q}{\partial u_{i,j}} \\
& = \sum_{t=1}^T r_j^{[t]} \, (1-r_j^{[t]}) \, r_i^{[t-1]} \, \frac{\partial Q}{\partial r_j^{[t]}}
+ \sum_{t=1}^T r_i^{[t-1]} \, h_j^{[t]},
\end{align}
where the last equality follows from \eqref{eq:DyBM:RTRBM:b}-\eqref{eq:DyBM:RTRBM:barh}.  In vector-matrix notations, we can write
\begin{align}
\nabla_{\mathbf{U}} Q
& = 
\sum_{t=1}^T 
\mathbf{r}^{[t-1]} \, 
\Big(\mathbf{r}^{[t]} \cdot (1-\mathbf{r}^{[t]}) \cdot \nabla_{\mathbf{r}^{[t]}} Q + \mathbf{h}^{[t]}\Big)^\top,
\end{align}
where $\nabla_{\mathbf{r}^{[t]}} Q$ is given by \eqref{eq:DyBM:RTRBM-Q}-\eqref{eq:DyBM:RTRBM-QT}.

The gradient of $Q$ with respect to other parameters can be derived as follows:
\begin{align}
\nabla_{\mathbf{W}} Q
& = 
\sum_{t=0}^T 
\mathbf{x}^{[t]} \, 
\Big(\mathbf{r}^{[t]} \cdot (1-\mathbf{r}^{[t]}) \cdot \nabla_{\mathbf{r}^{[t]}} Q\Big)^\top \\
\nabla_{\mathbf{b}^{\rm H}} Q
& = 
\sum_{t=1}^T 
\mathbf{r}^{[t]} \cdot (1-\mathbf{r}^{[t]}) \cdot \nabla_{\mathbf{r}^{[t]}} Q \\
\nabla_{\mathbf{b}^{\rm init}} Q
& = 
\mathbf{r}^{[0]} \cdot (1-\mathbf{r}^{[0]}) \cdot \nabla_{\mathbf{r}^{[0]}} Q \\
\nabla_{\mathbf{b}^{\rm V}} Q
& = \mathbf{0}
\end{align}

The gradients of $Q$ can be used to show the following gradients of the energy:
\begin{align}
\nabla_{\mathbf{U}} E_\theta(\mathbf{x},\mathbf{h})
& = 
-\sum_{t=1}^T 
\mathbf{r}^{[t-1]} \, 
\Big(\mathbf{r}^{[t]} \cdot (1-\mathbf{r}^{[t]}) \cdot \nabla_{\mathbf{r}^{[t]}} Q + \mathbf{h}^{[t]}\Big)^\top 
\\
\nabla_{\mathbf{W}} E_\theta(\mathbf{x},\mathbf{h})
& = 
-\sum_{t=0}^T \mathbf{x}^{[t]} \, (\mathbf{h}^{[t]})^\top
-\sum_{t=0}^T 
\mathbf{x}^{[t]} \, 
\Big(\mathbf{r}^{[t]} \cdot (1-\mathbf{r}^{[t]}) \cdot \nabla_{\mathbf{r}^{[t]}} Q\Big)^\top 
\\
\nabla_{\mathbf{b}^{\rm H}} E_\theta(\mathbf{x},\mathbf{h})
& = 
-\sum_{t=1}^T \mathbf{h}^{[t]}
-\sum_{t=1}^T 
\mathbf{r}^{[t]} \cdot (1-\mathbf{r}^{[t]}) \cdot \nabla_{\mathbf{r}^{[t]}} Q 
\\
\nabla_{\mathbf{b}^{\rm init}} E_\theta(\mathbf{x},\mathbf{h})
& = 
-\mathbf{h}^{[0]}
-\mathbf{r}^{[0]} \cdot (1-\mathbf{r}^{[0]}) \cdot \nabla_{\mathbf{r}^{[0]}} Q 
\\
\nabla_{\mathbf{b}^{\rm V}} E_\theta(\mathbf{x},\mathbf{h})
& = -\sum_{t=0}^T \mathbf{x}^{[t]},
\end{align}
where $\nabla_{\mathbf{r}^{[t]}} Q$ is given by \eqref{eq:DyBM:RTRBM-Q}-\eqref{eq:DyBM:RTRBM-QT}.  The gradient of the log-likelihood of the given time-series $\mathbf{x}$ now follows from \eqref{eq:BM:gradH-hidden-2} in \cite{BMsurvey1}.

\subsubsection{Extensions}

The RTRBM has been extended in various ways.  Mittleman et al.\ study a
structured RTRBM, where units are partitioned into blocks, and only the
connections between particular blocks are allowed \cite{SRTRBM}.  Lyu et
al.\ replaces the RNN of RTRBM with the one with Long Short-Term Memory
(LSTM) \cite{LSTM-RTRBM}.  Schrauwen and Buesing replaces the RNN of
RTRBM with an echo state network \cite{ESNRBM}.

An RNN-RBM slightly generalizes RTRBM by relaxing the constraint of the
RTRBM that $\mathbf{r}^{[t]}$ must be the expected value of
$\mathbf{h}^{[t]}$ \cite{RNNRBM}.  Namely, an RNN-RBM is an RNN but
gives an RBM as an output, where the RNN and RBM do not share
parameters, while an RTRBM shares parameters between an RNN and an RBM.

\section{Dynamic Boltzmann machines}
\label{sec:DyBM}

BPTT is not desirable for online learning, where we update $\theta$
every time a new pattern $\mathbf{x}^{[t]}$ is observed.  The per-step
computational complexity of BPTT in online learning grows linearly with
the length of the preceding time-series.  Such online learning, however,
is needed for example when we cannot store all observed patterns in
memory or when we want to adapt to changing environment.

The dynamic Boltzmann machine (DyBM) is proposed as a time-series model
that allows efficient online learning \cite{RT0967,DyBM}.  The per-step
computational complexity of the learning rule of a DyBM is independent
of the length of the preceding time-series.  In
Section~\ref{sec:DyBM:DyBM}, we start by reviewing the DyBM introduced
in \cite{RT0967,DyBM} with relation of its learning rule to spike-timing dependent plasticity (STDP).

In Section~\ref{sec:DyBM:flexibility}, we study the relaxation of 
some of the constraints that the DyBM has required in 
\cite{RT0967,DyBM} in a way that it becomes more suitable for 
inference and learning \cite{spike}.  The primary purpose of these constraints in 
\cite{RT0967,DyBM} was to mimic a particular form of STDP.  The
relaxed DyBM generalizes the original DyBM and allows us to interpret 
it as a form of logistic regression for time-series data.  

In Section~\ref{sec:DyBM:real}, we review DyBMs dealing with real-valued
time-series \cite{spike,NonlinearDyBM}.  These DyBMs are analogous to
how Gaussian Boltzmann machines \cite{GaussianBM,WRH04,HinSal06} deal
with real-valued patterns as opposed to Boltzmann machines
\cite{AHS85,HinSej83} for binary values.  The Gaussian DyBM can be
related to a vector autoregressive (VAR) model \cite{VAR}.  Specifically, we show
that a special case of the Gaussian DyBM is a VAR model having
additional variables that capture long term dependency of time-series.
These additional variables correspond to DyBM's eligibility traces,
which represent how recently and frequently spikes arrived from a neuron
to another.  We also review an extension of the Gaussian DyBM to deal
with time-series patterns in continuous space \cite{FunctionalDyBM}.

Some of the models and learning algorithms in this section have been
implemented in Python or Java and open-sourced at
\url{https://github.com/ibm-research-tokyo/dybm}.

\subsection{Dynamic Boltzmann machines for binary-valued time-series}
\label{sec:DyBM:DyBM}

\subsubsection{Finite dynamic Boltzmann machines\protect\footnote{This section closely follows \cite{RT0967}.}}

\begin{figure}
  \centering
 \includegraphics[width=0.3\linewidth]{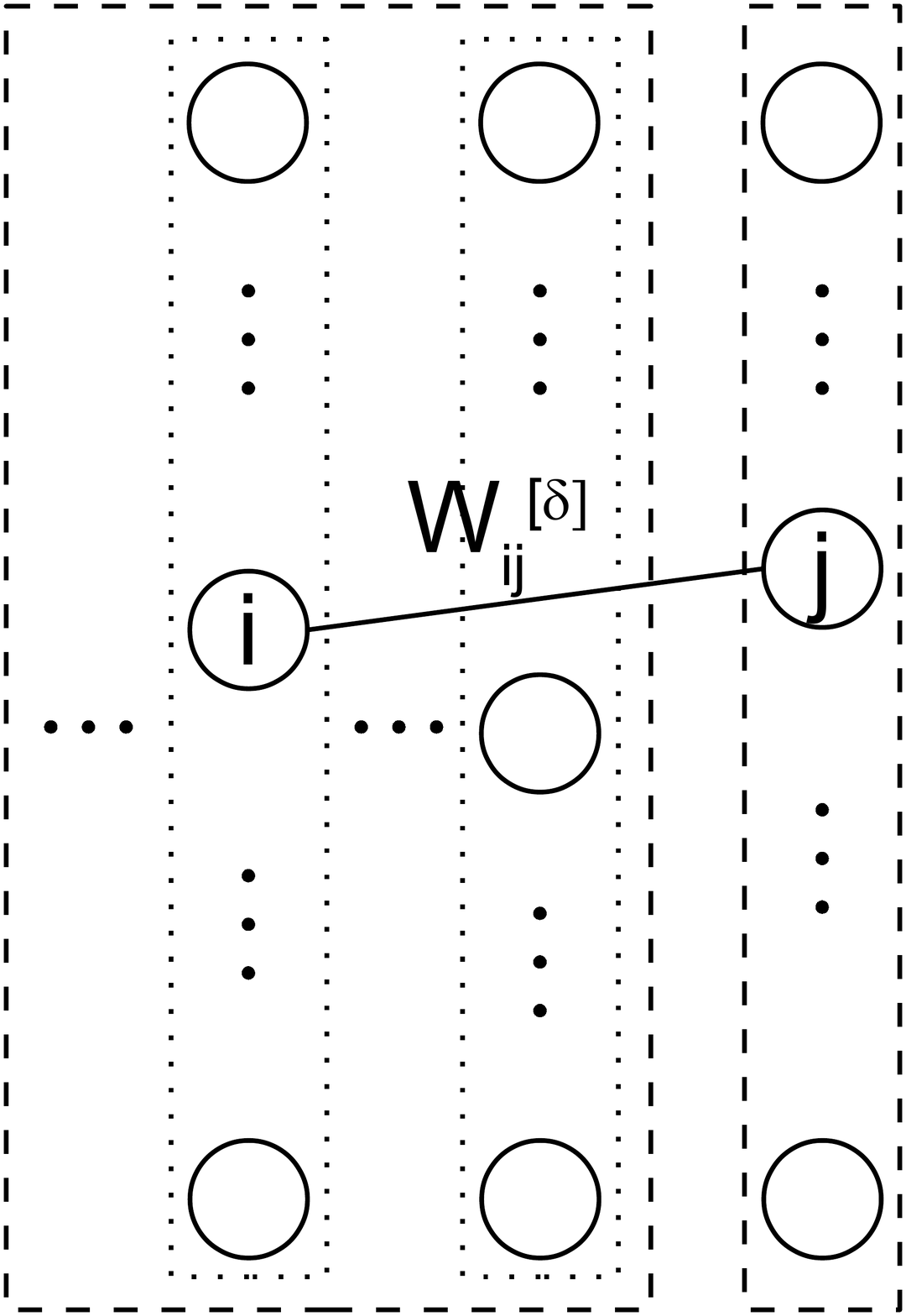}
 \caption{A dynamic Boltzmann machine unfolded through time (Figure~1(c) from \cite{RT0967}).}
 \label{fig:DyBM-unfolded}
 \end{figure}

The DyBM in \cite{RT0967,DyBM} is defined as a limit of a sequence of
Boltzmann machines (DyBM-$T$) consisting of $T$ layers as $T$ tends to
infinity (see Figure~\ref{fig:DyBM-unfolded}).  Formally, the DyBM-$T$ is
defined as the CRBM (see Section~\ref{sec:CRBM}) having $T$ layers of
$N$ visible units ($T-1$ layers of input units and one layer of output
units) and no hidden units, so that its conditional energy is defined as
\begin{align}
 E_\theta( \mathbf{x}^{[t]} \mid \mathbf{x}^{[t-T+1,t-1]} )
 & = - \mathbf{b}^\top \mathbf{x}^{[t]}
 - \sum_{\delta=1}^{T-1} (\mathbf{x}^{[t-\delta]})^\top \, \mathbf{W}^{[\delta]} \, \mathbf{x}^{[t]},
 \label{eq:dybmT}
\end{align}
where the weight of the DyBM-$T$ ($\mathbf{W}^{[1]}, \ldots,
\mathbf{W}^{[T-1]}$) assumes a particular parametric form with a finite
number of parameters that are independent of $T$, which we discuss in
the following.

The parametric form of the weight in the DyBM-$T$ is motivated by
observations from biological neural networks \cite{abbott2000synaptic}
but leads to particularly simple, exact, and efficient learning rule.
In biological neural networks, STDP has been postulated and supported
experimentally.  In particular, the weight from a pre-synaptic
neuron to a post-synaptic neuron is strengthened, if the post-synaptic
neuron fires (generates a spike) shortly {\em after} the pre-synaptic
neuron fires ({\it i.e.}, long term potentiation or LTP).  This weight is
weakened, if the post-synaptic neuron fires shortly {\em before} the
pre-synaptic neuron fires ({\it i.e.}, long term depression or LTD).  These
dependency on the timing of spikes is missing in the Hebbian rule for
the Boltzmann machine (see \eqref{eq:BM:update-w} from
\cite{BMsurvey1}).

\begin{figure}[t]
\centering \includegraphics[width=0.5\linewidth]{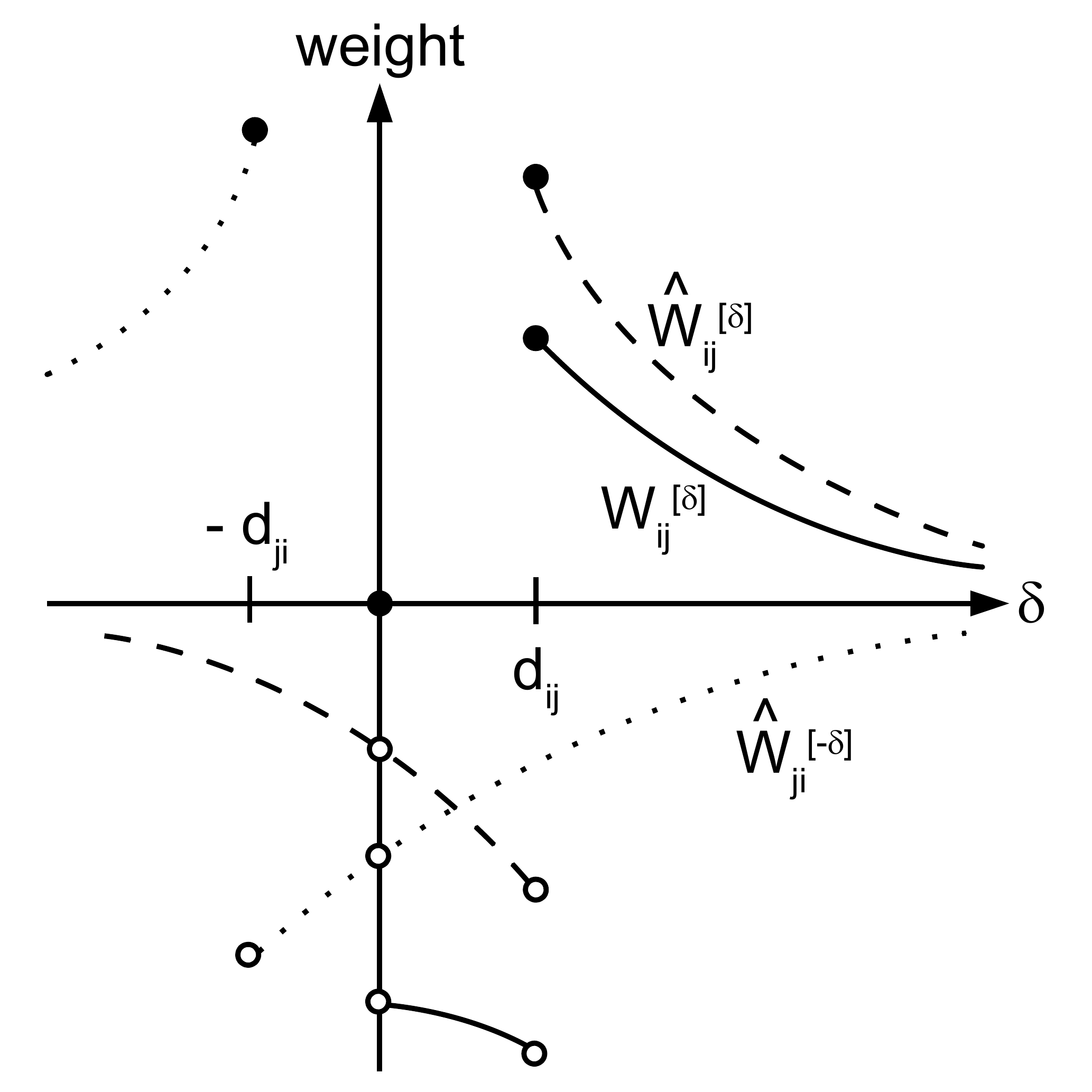}
 \caption{
The figure illustrates Equation~(\ref{eq:W}) with particular forms of
Equation~(\ref{eq:geometric}) (Figure~2 from \cite{RT0967}).  The horizontal axis represents $\delta$,
and the vertical axis represents the value of $w_{i,j}^{[\delta]}$ (solid
curves), $\hat w_{i,j}^{[\delta]}$ (dashed curves), or $\hat
w_{j,i}^{[-\delta]}$ (dotted curves).  Notice that $w_{i,j}^{[\delta]}$ is
defined for $\delta > 0$ and is discontinuous at $\delta = d$.  On the
other hand, $\hat w_{i,j}^{[\delta]}$ and $\hat w_{j,i}^{[-\delta]}$ are
defined for $-\infty < \delta < \infty$ and discontinuous at $\delta =
d_{i,j}$ and $\delta = -d_{j,i}$, respectively, where recall that we assume $d_{i,j}=d_{j,i}=d$ in this paper.}
\label{fig:weight}
\end{figure}

To have a learning rule with the characteristics of STDP with LTP and LTD, the DyBM-$T$ assumes
the weight of the form illustrated in
Figure~\ref{fig:weight}.  For $\delta>0$, we define the weight,
$w_{i,j}^{[\delta]}$, as the sum of two weights, $\hat
w_{i,j}^{[\delta]}$ and $w_{j,i}^{[-\delta]}$:
\begin{align}
w_{i,j}^{[\delta]} & = \hat w_{i,j}^{[\delta]} + \hat w_{j,i}^{[-\delta]},
\label{eq:W}
\intertext{where}
\hat w_{i,j}^{[\delta]} & = \left\{\begin{array}{ll}
0 & \mbox{if } \delta = 0\\
u_{i,j} \, \lambda^{\delta-d} & \mbox{if } \delta \ge d\\
-v_{i,j} \, \mu^{-\delta} & \mbox{otherwise}.
\end{array}\right.
\label{eq:geometric}
\end{align}
for $\lambda,\mu\in[0,1)$.  For simplicity, we assume a single decay
rate $\lambda$ for $\delta\ge d$ and a single decay rate $\mu$ for
$\delta<d$, as opposed to multiple ones in \cite{RT0967,DyBM}.
For simplicity, we assume that the conduction delay $d$ is uniform for all
connections, as opposed to variable conduction delay in \cite{DyBM}.
See also \cite{DelayPruning,HyperDyBM} for ways to tune the values of
the conduction delay.

In Figure~\ref{fig:weight},
the value of $\hat w_{i,j}^{[\delta]}$ is high when $\delta=d$,
the conduction delay from $i$-th (pre-synaptic) unit to the $j$-th
(post-synaptic) unit.  Namely, the
post-synaptic neuron is likely to fire ({\it i.e.}, $x_j^{[0]}=1$)
immediately after the spike from the pre-synaptic unit arrives with
the delay of $d$ ({\it i.e.}, $x_i^{[-d]}=1$).  This likelihood is
controlled by the LTP weight $u_{i,j}$.
The value of $\hat w_{i,j}^{[\delta]}$ gradually
decreases, as $\delta$ increases from $d$.  That is, the effect
of the stimulus of the spike arrived from the $i$-th unit diminishes
with time \cite{abbott2000synaptic}.

The value of $\hat w_{i,j}^{[d-1]}$ is low, suggesting that the
post-synaptic unit is unlikely to fire ({\it i.e.}, $x_j^{[0]}=1$) immediately
{\em before} the spike from the $i$-th (pre-synaptic) unit arrives.
This unlikelihood is controlled by the LTD weight $v_{i,j}$.  As
$\delta$ decreases from $d-1$, the magnitude of $\hat
w_{i,j}^{[\delta]}$ gradually decreases \cite{abbott2000synaptic}.
Here, $\delta$ can get smaller than 0, and $\hat w_{i,j}^{[\delta]}$
with $\delta<0$ represents the weight between the spike of the
pre-synaptic neuron that is generated after the spike of the
post-synaptic neuron.

\subsubsection{Dynamic Boltzmann machine as a limit of a sequence of finite dynamic Boltzmann machines\protect\footnote{This section closely follows \cite{spike}.}}
\label{sec:dybm:dybm}

The DyBM is defined as a limit of the sequence of DyBM-$T$ as
$T\to\infty$.  Because each DyBM-$T$ is a CRBM, we can also define the
limit of the sequence of the conditional probability defined by
DyBM-$T$, and this limit is considered as the conditional probability
defined by the DyBM.  Likewise, the conditional energy of the DyBM is defined as the
limit of the sequence of the conditional energy of DyBM-$T$.

Specifically, as $T\to\infty$, the conditional energy of DyBM-$T$ in
\eqref{eq:dybmT} converges to
\begin{align}
 E_\theta( \mathbf{x}^{[t]} \mid \mathbf{x}^{[<t]} )
 & = - \mathbf{b}^\top \mathbf{x}^{[t]}
 - \sum_{d=1}^{\infty} (\mathbf{x}^{[t-d]})^\top \, \mathbf{W}^{[d]} \, \mathbf{x}^{[t]},
 \label{eq:dybm:inf}
\end{align}
where the convergence is due to the parametric form
\eqref{eq:geometric}.  This conditional energy in turn defines the
conditional distribution via \eqref{eq:time-series:conditional}, where we
now have no hidden units.  Although the conditional energy
\eqref{eq:dybm:inf} of the DyBM involves an infinite sum, it can be
evaluated with a finite sum because of the parametric form
\eqref{eq:geometric}.

\begin{figure}[t]
 \centering
 \includegraphics[width=\linewidth]{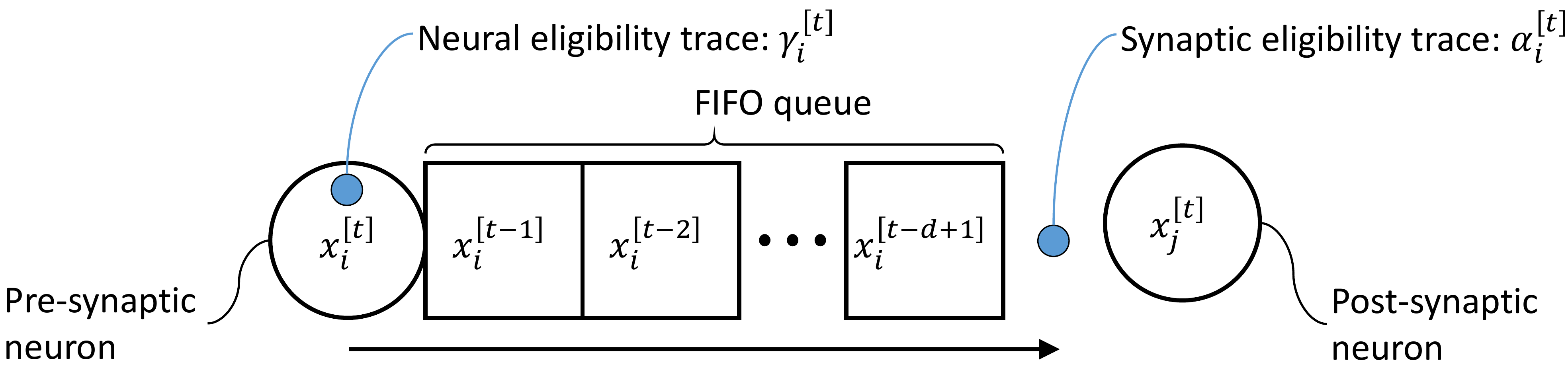}
 \caption{A connection from a (pre-synaptic) neuron $i$ to a (post-synaptic) neuron $j$ in a DyBM.}
 \label{fig:DyBM}
\end{figure}

In fact, the DyBM can be understood as an artificial model of a spiking
neural network where all computation for inference and learning is
performed locally at each synapse using only the information available
around the synapse.  Specifically, a (pre-synaptic) neuron is connected
to a (post-synaptic) neuron via a first-in-first-out (FIFO) queue and a
synapse (see Figure~\ref{fig:DyBM}).  At each discrete time $t$, a
neuron $i$ either fires ($x_i^{[t]}=1$) or not ($x_i^{[t]}=0$). The
spike travels along the FIFO queue and reaches the synapse after
conduction delay, $d$.  In other words, the
FIFO queue has the length of $d-1$ and stores, at time $t$, the spikes
that have been generated by the pre-synaptic neuron from time $t-d+1$ to
time $t-1$.

\begin{figure}[t]
 \centering
 \includegraphics[width=0.5\linewidth]{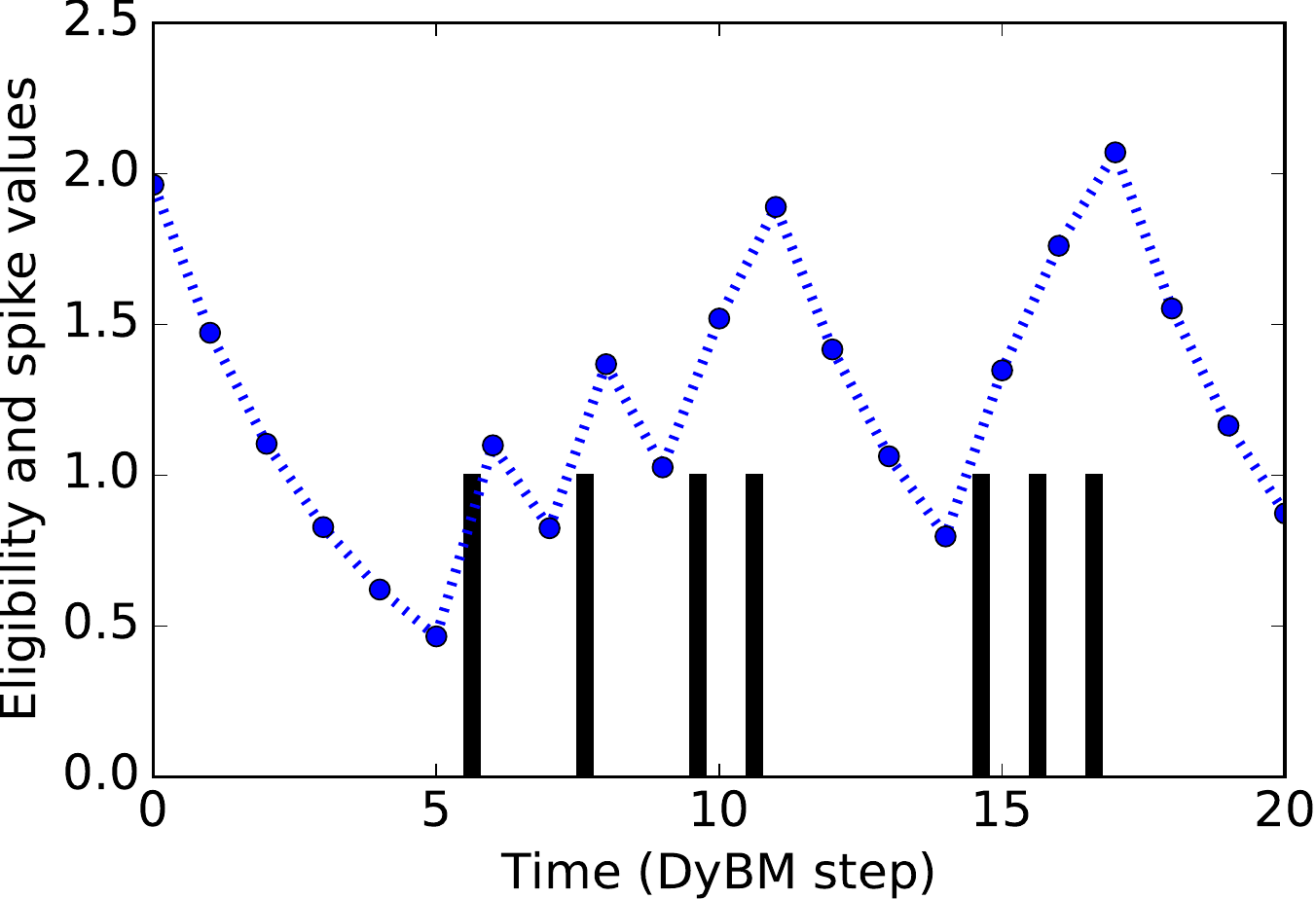}
 \caption{The value of a synaptic or neural eligibility trace as a function of time.
 For a synaptic eligibility trace at a synapse, the bars represent the spikes arrived from a FIFO queue at that synapse.
 For a neural eligibility trace at a neuron, the bars represent the 
 spikes generated by that neuron.}
 \label{fig:etrace}
\end{figure}

Each synapse in a DyBM stores a quantity called a synaptic 
eligibility trace.  The value of the 
synaptic eligibility increases when a spike arrives at the synapse from the FIFO 
queue; otherwise, it is decreased by a constant factor.  
Specifically, at time $t$, the value of the synaptic eligibility trace,
$\alpha_{i}^{[t]}$, that is stored at the synapse from a pre-synaptic neuron $i$ is updated as 
follows:
\begin{align}
 \alpha_{i}^{[t]} = \lambda \, (\alpha_{i}^{[t-1]} + x_i^{[t-d+1]}),
 \label{eq:synaptic}
\end{align} 
where $\lambda$ is a decay rate and satisfies $0\le\lambda<1$.  
Figure~\ref{fig:etrace} shows an example of how the value of the synaptic 
eligibility trace changes depending on the spikes arrived at the synapse. 
 Observe that $\alpha_{i}^{[t]}$ represents how recently and 
frequently spikes arrived from a pre-synaptic neuron $i$ and can be represented non-recursively as follows:
\begin{align}
 \alpha_i^{[t-1]} = \sum_{s=-\infty}^{t-d} \lambda^{t-s-d} \, x_i^{[s]}.
 \label{eq:DyBM:etrace}
\end{align}

Each neuron in a DyBM stores a quantity called a neural eligibility 
trace\footnote{We assume a single neural eligibility trace, as opposed 
to multiple ones in \cite{DyBM}, at each neuron.}. 
The value of the neural eligibility increases when the neuron fires; 
otherwise, it is decreased by a constant factor.  Specifically, at time 
$t$, the value of the neural eligibility trace, $\gamma_{i}^{[t]}$, at 
a neuron $i$ is updated as follows:
\begin{align}
 \gamma_{i}^{[t]} = \mu \, (\gamma_{i}^{[t-1]} + x_i^{[t]}),
 \label{eq:neural}
\end{align} 
where $\mu$ is a decay rate and satisfies $0\le\mu<1$.  
Observe that $\gamma_{i}^{[t]}$ represents how recently and 
frequently the neuron $i$ has fired and can be represented non-recursively as follows:
\begin{align}
 \gamma_{i}^{[t-1]} = \sum_{s=-\infty}^{t-1} \mu^{t-s} \, x_j^{[s]}
\end{align} 

A neuron in a DyBM fires according to the probability distribution that 
depends on the energy of the DyBM.  A neuron is more likely to fire when 
the energy becomes lower if it fires than otherwise.  
Let $E_{\theta,j}\big(x_j^{[t]}|\mathbf{x}^{[<t]}\big)$ be
the energy associated with a neuron $j$ at time $t$, which can depend on whether $j$ 
fires at time $t$ ({\it i.e.}, $x_j^{[t]}$) as well as the preceding spiking activities 
of the neurons in the DyBM ({\it i.e.}, $\mathbf{x}^{[<t]}$).  The firing probability of 
a neuron $j$ is then given by
\begin{align}
\PP_{\theta,j}( x_j^{[t]} | \mathbf{x}^{[<t]} )
& = 
\frac{\exp\big( -E_{\theta,j}(x_j^{[t]}|\mathbf{x}^{[<t]}) \big)}
{\displaystyle\sum_{\tilde x\in\{0,1\}} \exp\big(-E_{\theta,j}(\tilde x|\mathbf{x}^{[<t]})\big)}
\label{eq:probability}
\end{align}
for $x_j^{[t]}\in\{0,1\}$.
Specifically, $E_{\theta,j}\big(x_j^{[t]}|\mathbf{x}^{[<t]}\big)$ can be represented as follows:
\begin{align} 
E_{\theta,j}\big(x_j^{[t]}|\mathbf{x}^{[<t]}\big)
=
- b_j \, x_j^{[t]}
+ E_{\theta,j}^{\rm LTP} \big(x_j^{[t]}|\mathbf{x}^{[<t]}\big)
+ E_{\theta,j}^{\rm LTD} \big(x_j^{[t]}|\mathbf{x}^{[<t]}\big),
\label{eq:energy}
\end{align} 
where $b_j$ is the bias parameter of 
a neuron $j$ and represents how likely $j$ spikes ($j$ is more likely to 
fire if $b_j$ has a large positive value), and we define
\begin{align}
E_{\theta,j}^{\rm LTP} \big(x_j^{[t]}|\mathbf{x}^{[<t]}\big)
& \equiv
- \sum_{i=1}^N u_{i,j} \, \alpha_{i}^{[t-1]} \, x_j^{[t]} \label{eq:LTPenergy}\\
E_{\theta,j}^{\rm LTD} \big(x_j^{[t]}|\mathbf{x}^{[<t]}\big)
& \equiv 
\sum_{i=1}^N v_{i,j} \, \beta_{i}^{[t-1]} \, x_j^{[t]}
+ \sum_{k=1}^N v_{j,k} \, \gamma_{k}^{[t-1]} \, x_j^{[t]}, \label{eq:LTDenergy}
\end{align}
where $\beta_{i}^{[t-1]}$ represents how soon and frequently spikes will arrive
at the synapse from the FIFO queues from $i$ to $j$:
\begin{align}
\beta_{i}^{[t-1]} \equiv \sum_{s=t-d+1}^{t-1} \mu^{s-t} \, x_i^{[s]}.
\label{eq:beta}
\end{align}
Although $\beta_i^{[t-1]}$ can also be represented in a recursive manner, recursively computed $\beta_i^{[t-1]}$ is prone to numerical instability.

In \eqref{eq:LTPenergy}, the summation with respect to $i$ is over all 
of the pre-synaptic neurons that are connected to $j$. Here, $u_{i,j}$ 
is the weight parameter from $i$ to $j$ and represents the strength of 
Long Term Potentiation (LTP).  This weight parameter is thus referred to 
as LTP weight.  A neuron $j$ is more likely to fire ($x_j^{[t]}=1$) when 
$\alpha_{i}^{[t-1]}$ is large for a pre-synaptic neuron $i$ connected to 
$j$ (spikes have recently arrived at $j$ from $i$) and the 
corresponding $u_{i,j}$ is positive and large (LTP from $i$ to $j$ is 
strong).

In \eqref{eq:LTDenergy}, the summation with respect to $i$ is over all 
of the pre-synaptic neurons that are connected to $j$, and the summation 
with respect to $k$ is over all of the post-synaptic neurons which $j$ 
is connected to. Here, $v_{i,j}$ represents the strength of Long Term 
Depression from $i$ to $j$ and referred to as LTD weight.  The neuron 
$j$ is less likely to fire when $\beta_{i}$ is large for a pre-synaptic 
neuron $i$ connected to $j$ (spikes will soon and frequently reach $j$ 
from $i$) and the corresponding $v_{i,j}$ is positive and large (LTD 
from $i$ to $j$ is strong).  The second term in \eqref{eq:LTDenergy} 
represents that a pre-synaptic neuron $j$ is less likely to fire if a 
post-synaptic neuron has recently and frequently fired ($\gamma_k$ is 
large), and the strength of this LTD is given by $v_{j,k}$.  Notice that 
the timing of a spike is measured with respect to when the spike reaches 
synapse, where the spike from a pre-synaptic neuron has the delay $d$, 
and the spike from a post-synaptic neuron reaches immediately.

The learning rule of the DyBM has been derived in a way that it maximizes the log likelihood of
given time-series with respect to the probability distribution given by \eqref{eq:probability} \cite{DyBM}.
Specifically, at time $t$, the DyBM updates its parameters according to 
\begin{align}
b_j & \leftarrow b_j + \eta \, \big( x_j^{[t]} - \E_{\theta,j}[ X_j^{[t]} \mid \mathbf{x}^{[<t]} ] \big) \label{eq:b} \\
u_{i,j} & \leftarrow u_{i,j} + \eta \, \alpha_{i}^{[t-1]} \, \big( x_j^{[t]} - \E_{\theta,j}[ X_j^{[t]} \mid \mathbf{x}^{[<t]} ] \big) \label{eq:u} \\
v_{i,j} & \leftarrow v_{i,j} + \eta \, \beta_{i}^{[t-1]} \, \big( \E_{\theta,j}[ X_j^{[t]} \mid \mathbf{x}^{[<t]} ] - x_j^{[t]} \big)
+ \eta \, \gamma_{j}^{[t-1]} \, \big( \langle X_i^{[t]} \rangle - x_i^{[t]} \big) \label{eq:v}
\end{align}
for each of neurons $i$ and $j$, where $\eta$ is a learning rate,
$x_j^{[t]}$ is the training data given to $j$ at time $t$, and
$\E_{\theta,j}[ X_j^{[t]} \mid \mathbf{x}^{[<t]} ]$ denotes the
expected value of $x_j^{[t]}$ ({\it i.e.}, firing probability of a
neuron $j$ at time $t$) according to the probability distribution given
by \eqref{eq:probability}.  By following stochastic gradient methods
\cite{SGD,Adam,AdaGrad,RMSProp,momentum}, the learning rate $\eta$ may
be adjusted over time $t$.

\subsubsection{Relation to spike-timing dependent plasticity\protect\footnote{This section closely follows \cite{spike}.}}

In spike-timing dependent plasticity (STDP),
the amount of the change in the weight between two neurons 
that fired together depends on the precise timings when the two neurons 
fired.  STDP supplements the Hebbian rule \cite{Hebb} and has been 
experimentally confirmed in biological neural networks \cite{BiPoo98}.  

In \eqref{eq:u}, $u_{i,j}$ is increased (LTP gets stronger) when 
$x_j^{[t]}=1$ is given to $j$.  Then $j$ becomes more likely to fire 
when spikes from $i$ have recently and frequently arrived at $j$ ({\it 
i.e.}, $\alpha_{i}^{[\cdot]}$ is large).  This amount of the change in 
$u_{i,j}$ depends on $\alpha_{i}^{[t-1]}$, exhibiting a key property of 
STDP.  In particular, $u_{i,j}$ is increased by a large amount if spikes 
from $i$ have recently and frequently arrived at $j$.  

According to the second term on the right-hand side of \eqref{eq:v}, 
$v_{i,j}$ is increased (LTD gets stronger) when $x_j^{[t]}=0$ is given 
to a post-synaptic neuron $j$.  Then $j$ becomes less likely to fire 
when spikes from $i$ are expected to reach $j$ soon ({\it i.e.}, 
$\beta_{i}^{[\cdot]}$ is large).  This amount of the change in $v_{i,j}$ 
is large if there are spikes in the FIFO queue from $i$ to $j$ and they 
are close to $j$. According to the last term of \eqref{eq:v}, $v_{i,j}$ 
is increased when $x_i^{[t]}=0$ is given to the pre-synaptic $i$, and 
this amount of the change in $v_{i,j}$ is proportional to $\gamma_j$ 
({\it i.e.}, how frequently and recently the post-synaptic $j$ has fired). 
This learning rule of \eqref{eq:v} thus exhibits some of the key properties of LTD 
with STDP.

In \eqref{eq:b}, $b_j$ is increased when $x_j^{[t]}=1$ is given to $j$,
so that $j$ becomes more likely to fire (in accordance with the training
data), but the amount of the change in $b_j$ is small if $j$ is already
likely to fire ($\E_{\theta,j}[ X_j^{[t]} \mid \mathbf{x}^{[<t]}
]\approx 1$). This dependency on $\E_{\theta,j}[ X_j^{[t]} \mid
\mathbf{x}^{[<t]} ]$ can be considered as a form of homeostatic
plasticity \cite{TurNel04,LPT09}.

\paragraph{Related work}

There has been a significant amount of the prior work towards
understanding STDP from the perspectives of machine learning \cite{NPBM13,BMFZW16,SceBen16}.  For
example, Nessler et al.\ show that STDP can be understood as
approximating the expectation maximization (EM) algorithm
\cite{NPBM13}.  Nessler et al.\ study a particularly structured
(winner-take-all) network and its learning rule for maximizing
the log likelihood of given static patterns.  On the other hand, the
DyBM does not assume particular structures in the
network, and the learning rule having the properties of STDP applies
for any synapse in the network.  Also, the learning rule of the DyBM
maximizes the log likelihood of given
time-series, and its learning rule does not involve approximations
beyond what is assumed in stochastic gradient methods \cite{SGD}.

\subsection{Giving flexibility to the DyBM\protect\footnote{This section closely follows \cite{spike}.}}
\label{sec:DyBM:flexibility}

It has been shown in \cite{DyBM} that the DyBM in 
Section~\ref{sec:DyBM:DyBM} has the capability of associative memory and 
anomaly detection for sequential patterns, but the applications of the 
DyBM have been limited to simple tasks with relatively low dimensional 
time-series.  In \cite{spike}, we relax some of the constraints of this DyBM in a 
way that it gives more flexibility that is useful for learning and inference.

Specifically, observe that the first term on the right-hand side of \eqref{eq:LTDenergy} can be
rewritten with the definition of $\beta_{i}^{[t-1]}$ in \eqref{eq:beta} as follows:
\begin{align}
\sum_{i=1}^N v_{i,j} \, \beta_{i}^{[t-1]} \, x_j^{[t]}
& = \sum_{i=1}^N \sum_{s=t-d+1}^{t-1} v_{i,j} \, \mu^{s-t} \, x_i^{[s]} \, x_j^{[t]} \\
& = \sum_{i=1}^N \sum_{\delta=1}^{d-1} v_{i,j}^{[\delta]} \, x_i^{[t-\delta]} \, x_j^{[t]},
\end{align}
where we let $v_{i,j}^{[\delta]} \equiv v_{i,j} \, \mu^{-\delta}$.  
Here, $v_{i,j}^{[\delta]}$ represents how unlikely $j$ fires at time $t$ 
if $i$ fired at time $t-\delta$.  The parametric form of 
$v_{i,j}^{[\delta]} \equiv v_{i,j} \, \mu^{-\delta}$ assumes that
this LTD weight decays geometrically as the interval, $\delta$, between the two spikes increases.

In the following, we relax this constraint on $v_{i,j}^{[\delta]}$ for 
$\delta=1,\ldots,d-1$ and assumes that these LTD weights can take 
independent values.  Then the energy of the DyBM with $N$ neurons
can be represented conveniently with matrix and vector operations:
\begin{align} 
E_\theta(\mathbf{x}^{[t]}|\mathbf{x}^{[<t]})
& \equiv \sum_{j=1}^N E_{\theta,j}(x_j^{[t]}|\mathbf{x}^{[<t]}) \\
& 
- \mathbf{b}^\top \mathbf{x}^{[t]}
- (\boldsymbol{\alpha}_\lambda^{[t-1]})^\top \mathbf{U} \, \mathbf{x}^{[t]}
+ \sum_{\delta=1}^{d-1} (\mathbf{x}^{[t-\delta]})^\top \mathbf{V}^{[\delta]} \, \mathbf{x}^{[t]}
+ (\mathbf{x}^{[t]})^\top \mathbf{V} \, \boldsymbol{\gamma}_\mu^{[t-1]},
\label{eq:matrix}
\end{align}
where $\mathbf{b}\equiv(b_j)_{j=1,\ldots,N}$ is a 
vector, $\mathbf{U}\equiv(u_{i,j})_{(i,j)\in\{1,\ldots,N\}^2}$ is a 
matrix, and other boldface letters are defined analogously (a vector is 
lowercase and a matrix is uppercase).  
For eligibility traces ($\boldsymbol{\alpha}_\lambda^{[t-1]}$ and
$\boldsymbol{\gamma}_\mu^{[t-1]}$), we append the subscript to explicitly
represent the dependency on the decay rate ($\lambda$ and $\mu$). 
The functional form of the energy 
completely determines the dynamics of a DyBM, and relaxing its constraints 
allows the DyBM to represent a wider class of dynamical 
systems.

Notice that the last term of \eqref{eq:matrix} can be divided into two terms:
\begin{align}
(\mathbf{x}^{[t]})^\top \mathbf{V} \, \boldsymbol{\gamma}_\mu^{[t-1]}
& = (\boldsymbol{\gamma}_\mu^{[t-1]})^\top \mathbf{V} \, \mathbf{x}^{[t]} \\
& = (\boldsymbol{\alpha}_\mu^{[t-1]})^\top \mathbf{V} \, \mathbf{x}^{[t]}
+ \sum_{\delta=1}^{d-1} (\mathbf{x}^{[t-\delta]})^\top \mathbf{\hat V}^{[\delta]} \, \mathbf{x}^{[t]},
\label{eq:divide}
\end{align}
where $\boldsymbol{\alpha}_\mu^{[t-1]}$ is the same as the vector of 
synaptic eligibility traces but with the decay rate $\mu$, and 
$\mathbf{\hat V}^{[\delta]} \equiv \mu^{-\delta} \, \mathbf{V}$.
Comparing \eqref{eq:divide} and \eqref{eq:matrix}, we find that, without loss of generality, 
the energy of the DyBM
can be represented with the following form:
\begin{align} 
E_\theta(\mathbf{x}^{[t]}|\mathbf{x}^{[<t]})
& =
- \bigg(
\mathbf{b}^\top
+ \sum_{\delta=1}^{d-1} (\mathbf{x}^{[t-\delta]})^\top \mathbf{W}^{[\delta]}
+ \sum_{\ell=1}^L (\boldsymbol{\alpha}_{\lambda_\ell}^{[t-1]})^\top \mathbf{U}^{[\ell]}
\bigg) \, \mathbf{x}^{[t]},
\label{eq:general}
\end{align} 
where we define $\mathbf{W}^{[\delta]} = -\mathbf{V}^{[\delta]} - \mathbf{\hat V}^{[\delta]}$.
The energy in \eqref{eq:general} reduces to the original energy in \eqref{eq:energy} when
$\mathbf{W}^{[\delta]} = - \mu^{-\delta} \, \mathbf{V} - \mu^{\delta} \, \mathbf{V}^\top$,
$\mathbf{U}^{[1]} = \mathbf{U}$,
$\mathbf{U}^{[2]} = - \mu^d \, \mathbf{V}^\top$,
$\lambda_1    = \lambda$, 
$\lambda_2    = \mu$,
and $L=2$.  With $L>2$, one can also incorporate multiple synaptic or neural eligibility
traces with varying decay rates in \cite{DyBM}. 

Equivalently, we can represent the energy using neural eligibility traces, 
$\boldsymbol{\gamma}_{\mu_\ell}$, instead of synaptic eligibility traces,
$\boldsymbol{\alpha}_{\lambda_\ell}$, as follows:
\begin{align} 
E_\theta(\mathbf{x}^{[t]}|\mathbf{x}^{[<t]})
& =
- \bigg(
\mathbf{b}^\top
+ \sum_{\delta=1}^{d-1} (\mathbf{x}^{[t-\delta]})^\top \mathbf{W}^{[\delta]}
+ \sum_{\ell=1}^L (\boldsymbol{\gamma}_{\mu_\ell}^{[t-1]})^\top \mathbf{V}_\ell
\bigg) \, \mathbf{x}^{[t]}.
\label{eq:general2}
\end{align} 

\subsubsection{Learning rule in vector-matrix notations}

The learning rule corresponding to the representation with \eqref{eq:general} is as follows:
\begin{align}
 \mathbf{b}
 & \leftarrow \mathbf{b} + \eta \, (\mathbf{x}^{[t]}-\E_\theta[\boldsymbol{X}^{[t]}\mid\mathbf{x}^{[<t]}]) \\
 \mathbf{W}^{[\delta]}
 & \leftarrow \mathbf{W}^{[\delta]} + \eta \, \mathbf{x}^{[t-\delta]} \, (\mathbf{x}^{[t]}-\E_\theta[\boldsymbol{X}^{[t]}\mid\mathbf{x}^{[<t]}] )^\top \\
 \mathbf{U}^{[\ell]}
 & \leftarrow \mathbf{U}^{[\ell]} + \eta \, \boldsymbol{\alpha}_{\lambda_{\ell}}^{[t-1]} \, (\mathbf{x}^{[t]}-\E_\theta[\boldsymbol{X}^{[t]}\mid\mathbf{x}^{[<t]}] )^\top
\end{align}
for each $\delta$ and each $\ell$, where $\E_\theta[\boldsymbol{X}^{[t]}\mid\mathbf{x}^{[<t]}]$ is the conditional expectation with respect to
\begin{align}
 \PP_\theta( \mathbf{x}^{[t]} \mid \mathbf{x}^{[<t]} )
 & = \frac{\exp(-E_\theta( \mathbf{x}^{[t]} \mid \mathbf{x}^{[<t]} ))}
 {\displaystyle\sum_{\mathbf{\tilde x}^{[t]}} \exp(-E_\theta( \mathbf{\tilde x}^{[t]} \mid \mathbf{x}^{[<t]} ))}.
\end{align}
Specifically,
\begin{align}
\E_{\theta}[\boldsymbol{X}^{[t]} \mid \mathbf{x}^{[<t]}]
& = \frac{\exp(\mathbf{m}^{[t]})}{1 + \exp(\mathbf{m}^{[t]})}
\label{eq:logit}
\end{align}
with
\begin{align}
\boldsymbol{m}^{[t]}
& \equiv \mathbf{b}^\top
+ \sum_{\delta=1}^{d-1} (\mathbf{x}^{[t-\delta]})^\top \mathbf{W}^{[\delta]}
+ \sum_{\ell=1}^L (\boldsymbol{\alpha}_{\lambda_\ell}^{[t-1]})^\top \mathbf{U}^{[\ell]},
\label{eq:mean}
\end{align}
where exponentiation and division of vectors are elementwise.

The form of \eqref{eq:logit} implies that the DyBM is a kind of a logit 
model, where the feature vector, $(\mathbf{x}^{[t-d+1]}, \ldots, 
\mathbf{x}^{[t-1]}, \boldsymbol{\alpha}_\lambda^{[t-1]}, 
\boldsymbol{\alpha}_\mu^{[t-1]})$, depends on the prior values, 
$\mathbf{x}^{[<t]}$, of the time-series. By applying the learning 
rules given in \eqref{eq:b}-\eqref{eq:v} to given time-series, we can 
learn the parameters of the DyBM or equivalently the parameters of the 
logit model ({\it i.e.}, $\mathbf{b}$, $\mathbf{W}^{[\delta]}$ for 
$\delta=1, \ldots, d-1$, and $\mathbf{U}^{[\ell]}$ for $\ell=1, \ldots, L$) in \eqref{eq:logit}.

\subsection{Dynamic Boltzmann machines for real-valued time-series}
\label{sec:DyBM:real}

\subsubsection{Gaussian dynamic Boltzmann machines\protect\footnote{This section closely follows \cite{spike}.}}
\label{sec:gaussianDyBM}

In this section, we show how a DyBM can deal with real-valued
time-series in the form of a Gaussian DyBM \cite{spike,NonlinearDyBM}.
A Gaussian DyBM assumes that $x_j^{[t]}$ follows a Gaussian distribution
for each $j$:
\begin{align}
 p_\theta^{(j)}(x_j^{[t]} | \mathbf{x}^{[<t]})
 = \frac{1}{\sqrt{2\,\pi\,\sigma_j^2}}
 \exp\Big(-\frac{\big(x_j^{[t]}-m_j^{[t]}\big)^2}{2\,\sigma_j^2}\Big),
 \label{eq:spike_prob}
\end{align}
where $m_j^{[t]}$ is given by \eqref{eq:mean}, and $\sigma_j^2$ is a 
variance parameter.  This Gaussian distribution 
is in contrast to the Bernoulli distribution of the DyBM 
given by \eqref{eq:probability}.

The conditional energy of the Gaussian DyBM can be represented as follows:
\begin{align}
 E_\theta( \mathbf{x}^{[t]} \mid \mathbf{x}^{[<t]} )
 & = \sum_{j=1}^N \frac{(x_j^{[t]}-m_j^{[t]})^2}{2\,\sigma_j^2} \\
 & = \sum_{j=1}^N \frac{(x_j^{[t]}-b_j)^2}{2\,\sigma_j^2}
 - \sum_{\delta=1}^{d-1} \sum_{i=1}^N \sum_{j=1}^N x_i^{[t-\delta]} \, w_{i,j}^{[\delta]} \, x_j^{[t]}
 - \sum_{\ell=1}^{L} \sum_{i=1}^N \sum_{j=1}^N \alpha_{i,\lambda_\ell}^{[t-1]} \, u_{i,j}^{[\ell]} \, x_j^{[t]} + C,
\end{align}
where $C$ is the term that does not depend on $\mathbf{x}^{[t]}$.
Because $C$ is canceled out between the numerator and the denominator
in \eqref{eq:time-series:conditional}, we omit it from the conditional
energy.  By letting $\mathbf{W}_\sigma^{[\delta]}$ be the matrix whose
$(i,j))$ element is $w_{i,j}^{[\delta]}/\sigma_j^2$ and
$\mathbf{U}_\sigma^{[\ell]}$ be the matrix whose $(i,j))$ element is
$u_{i,j}^{[\ell]}/\sigma_j^2$, the conditional energy of the Gaussian
DyBM can be represented as follows:
\begin{align}
 E_\theta( \mathbf{x}^{[t]} \mid \mathbf{x}^{[<t]} )
 & = \sum_{j=1}^N \frac{(x_j^{[t]}-b_j)^2}{2\,\sigma_j^2}
 - \sum_{\delta=1}^{d-1} (\mathbf{x}^{[t-\delta]})^\top \, \mathbf{W}_{\sigma}^{[\delta]} \, \mathbf{x}^{[t]}
 - \sum_{\ell=1}^{L} (\mathbf{\alpha}_{\lambda_\ell}^{[t-1]})^\top \, \mathbf{U}_\sigma^{[\ell]} \, \mathbf{x}^{[t]}.
\end{align}
The conditional energy of the Gaussian DyBM may be compared against the
energy of the Gaussian Bernoulli restricted Boltzmann machine (see
\cite{GBRBM} or \eqref{eq:GBRBM-energy} from \cite{BMsurvey1}).

We now derive a learning rule for the Gaussian DyBM in a way that it 
maximizes the log-likelihood of given time-series $\mathbf{x}$:
\begin{align}
 \sum_t \log p_\theta(\mathbf{x}^{[t]}|\mathbf{x}^{[<t]})
& = \sum_t \sum_{i=1}^N \log p_i(x_i^{[t]}|\mathbf{x}^{[-\infty,t-1]}),
\label{eq:independence}
\end{align}
where the summation over $t$ is over all of the time steps of 
$\mathbf{x}$, and the conditional independence between 
$x_i^{[t]}$ and $x_j^{[t]}$ for $i\neq j$ given $\mathbf{x}^{[<t]}$ is 
the fundamental property of the DyBM as shown in \cite{DyBM}.

The approach of stochastic gradient is to update the parameters of the 
Gaussian DyBM at each step, $t$, according to the gradient of the 
conditional probability density of $\mathbf{x}^{[t]}$:
\begin{align}
 \nabla \log p_\theta(\mathbf{x}^{[t]}|\mathbf{x}^{[<t]})
& = - \sum_{i=1}^N
 \Big(\frac{1}{2} \nabla \log \sigma_i^2 + \nabla\frac{\big(x_i^{[t]}-m_i^{[t]})^2}{2\,\sigma_i^2}\Big),
\label{eq:gradient}
\end{align}
where the equality follow from
\eqref{eq:spike_prob}.
From \eqref{eq:gradient} and \eqref{eq:mean},
we can derive the derivative with respect to
each parameter as follows:
\begin{align}
 \frac{\partial}{\partial b_j} \log p_\theta(\mathbf{x}^{[t]}|\mathbf{x}^{[-\infty,t-1]})
 & = \frac{x_j^{[t]}-\mu_j^{[t]}}{\sigma_j^2} \, x_j^{[t]}
 \\
 \frac{\partial}{\partial u_{i,j}} \log p_\theta(\mathbf{x}^{[t]}|\mathbf{x}^{[-\infty,t-1]})
 & = \frac{x_j^{[t]}-\mu_j^{[t]}}{\sigma_j^2} \, \alpha_{i,j}^{[t-1]}
 \\
 \frac{\partial}{\partial w_{i,j}^{[\delta]}} \log p_\theta(\mathbf{x}^{[t]}|\mathbf{x}^{[-\infty,t-1]})
 & = \frac{x_j^{[t]}-\mu_j^{[t]}}{\sigma_j^2} \, x_i^{[t-\delta]}
 \\
 \frac{\partial}{\partial \sigma_j} \log p_\theta(\mathbf{x}^{[t]}|\mathbf{x}^{[-\infty,t-1]})
 & = - \frac{1}{\sigma_j} + \frac{\big(x_j^{[t]}-\mu_j^{[t]}\big)^2}{\sigma_j^3},
\end{align}
where $\delta\in\{1,\ldots,d-1\}$, $\ell\in\{1,\ldots,\}$, and
$i,j\in\{1,\ldots,N\}$.

These parameters are thus updated with learning rate $\eta$ as follows:
\begin{align}
 \mathbf{b}
 & \leftarrow \mathbf{b} + \eta \, \frac{\mathbf{x}^{[t]}-\mathbf{m}^{[t]}}{\boldsymbol{\sigma}^2}
 \label{eq:learning_rule1}\\
 \boldsymbol{\sigma}
 & \leftarrow \boldsymbol{\sigma} + \eta \, \frac{\big(\mathbf{x}^{[t]}-\mathbf{m}^{[t]}\big)^2 - \boldsymbol{\sigma}^2}{\boldsymbol{\sigma}^3}
 \\
 \mathbf{W}^{[\delta]}
 & \leftarrow \mathbf{W}^{[\delta]} + \eta \, \mathbf{x}^{[t-\delta]} \, \Bigg(\frac{\mathbf{x}^{[t]}-\mathbf{m}^{[t]}}{\boldsymbol{\sigma}^2} \Bigg)^\top \\
 \mathbf{U}^{[\ell]}
 & \leftarrow \mathbf{U}^{[\ell]} + \eta \, \boldsymbol{\alpha}_{\lambda_{\ell}}^{[t-1]} \, \Bigg(\frac{\mathbf{x}^{[t]}-\mathbf{m}^{[t]}}{\boldsymbol{\sigma}^2} \Bigg)^\top
\label{eq:learning_rule2}
\end{align}
where division and exponentiation of vectors are elementwise, and
$\mathbf{m}^{[t]}$ is given by \eqref{eq:mean}.

The maximum likelihood estimator of $\mathbf{x}^{[t]}$ by the Gaussian
DyBM is given by $\boldsymbol{m}^{[t]}$ in $\eqref{eq:mean}$.  The
Gaussian DyBM can thus be understood as a modification to the standard
vector autoregressive (VAR) model. Specifically, the last term in the
right-hand side of \eqref{eq:mean} involves eligibility traces, which
can be understood as features of historical values,
$\mathbf{x}^{(-\infty,t-d]}$, and are added as new variables to the VAR
model.  Because the value of the eligibility traces can depend on the
infinite past, the Gaussian DyBM can take into account the history
beyond the lag $d-1$.

\subsubsection{Natural gradients\protect\footnote{This section closely follows \cite{spike}.}}
\label{sec:natural}

In this section, we study a learning rule based on natural gradient for the Gaussian DyBM.
Consider a stochastic model that gives the probability density
of a pattern $\mathbf{x}$ as $p_\theta(\mathbf{x})$.
With natural gradients \cite{NaturalGradient}, the parameters,
$\theta$, of the stochastic model are updated as follows:
\begin{align}
 \theta_{t+1} = \theta_t - \eta \, G^{-1}(\theta_t) \, \nabla \log p_\theta(\mathbf{x})
 \label{eq:natural}
\end{align}
at each step $t$, where $\eta$ is the learning rate, and
$G(\theta)$ denotes the Fisher information matrix:
\begin{align}
 G(\theta) & \equiv \int p_\theta(\mathbf{x}) \left(\nabla \log p_\theta(\mathbf{x}) \, \nabla \log p_\theta(\mathbf{x})^\top \right) d\mathbf{x}.
\end{align}

Due to the conditional independence in \eqref{eq:independence}, it
suffices to
derive a natural gradient for each Gaussian unit.  Here, we consider the
parametrization with mean $m$ and variance $v\equiv\sigma^2$.  The
probability density function of a Gaussian distribution is represented
with this parametrization as follows:
\begin{align}
p(x;m,v) & = \frac{1}{\sqrt{2\pi\,v}} \exp\left(-\frac{(x-m)^2}{2v}\right).
\end{align}
The log likelihood of $x$ is then given by
\begin{align}
 \log p(x;m,v) & = -\frac{(x-m)^2}{2v} - \frac{1}{2} \log v - \frac{1}{2}\log 2\pi.
 \label{eq:loglikelihood}
\end{align}

Hence, the gradient and the inverse Fisher information matrix in \eqref{eq:natural} are given as follows:
\begin{align}
 \nabla \log p_\theta(\mathbf{x}) & = \left(\begin{array}{c}
		       \frac{x-m}{v}\\
			    \frac{(x-m)^2}{2v^2} - \frac{1}{2v}
					    \end{array}\right)\\
 G^{-1}(\theta)  &
 = \left(\begin{array}{cc}
    \frac{1}{v} & 0 \\
    0 & \frac{1}{2v^2}
   \end{array}\right)^{-1}
 = \left(\begin{array}{cc}
    v & 0 \\
    0 & 2v^2
   \end{array}\right),
\end{align}

The parameters $\theta_t\equiv(m_t,v_t)$ are then updated as follows:
\begin{align}
m_{t+1}
 & = m_t + \eta \, (x-m_t) \\
v_{t+1}
 & = v_t + \eta \, \left((x-m_t)^2 - v_t\right).
\end{align}

In the context of a Gaussian DyBM, the mean is given by \eqref{eq:mean},
where $m_j^{[t]}$ is linear with respect to $b_j$, $w_{i,j}$, and $u_{i,j}^{[\ell]}$.
Also, the variance is given by $\sigma_j^2$.  Hence, the natural gradient gives
the learning rules for these parameters as follows:
\begin{align}
 \mathbf{b}
 & \leftarrow \mathbf{b} + \eta \, (\mathbf{x}^{[t]}-\mathbf{m}^{[t]})
 \label{eq:natural_rule1}\\
 \boldsymbol{\sigma}^2
 & \leftarrow \boldsymbol{\sigma}^2 + \eta \, \Big(\big(\mathbf{x}^{[t]}-\mathbf{m}^{[t]}\big)^2 - \boldsymbol{\sigma}^2\Big) \\
 \mathbf{W}^{[\delta]}
 & \leftarrow \mathbf{W}^{[\delta]} + \eta \, \mathbf{x}^{[t-\delta]} \, (\mathbf{x}^{[t]}-\mathbf{m}^{[t]} )^\top \\
 \mathbf{U}^{[\ell]}
 & \leftarrow \mathbf{U}^{[\ell]} + \eta \, \boldsymbol{\alpha}_{\lambda_{\ell}}^{[t-1]} \, (\mathbf{x}^{[t]}-\mathbf{m}^{[t]} )^\top,
\label{eq:natural_rule2}
\end{align}
where the exponentiation of a vector is elementwise.  We can compare
\eqref{eq:natural_rule1}-\eqref{eq:natural_rule2} against what the
standard gradient gives in
\eqref{eq:learning_rule1}-\eqref{eq:learning_rule2}.

\subsubsection{Using nonlinear features in Gaussian DyBMs}

The Gaussian DyBM is a linear model and has limited capability in
modeling complex time-series.  A way to take into account non-linear
features of time-series with a Gaussian DyBM is to apply non-linear
mapping to input time-series and feed the resulting non-linear features
as additional input to the Gaussian DyBM.  An example of such non-linear
mapping is an echo state network (ESN)~\cite{ESN}.

An ESN maps an input sequence, $\mathbf{x}$, into $\boldsymbol{\psi}$ recursively as follows:
\begin{align}
\boldsymbol{\psi}^{[t]}
& = (1-\rho) \, \boldsymbol{\psi}^{[t-1]}
 + \rho \, \tanh\left( \mathbf{W}_{\rm rec} \, \boldsymbol{\psi}^{[t-1]} + \mathbf{W}_{\rm in} \, \mathbf{x}^{[t]} \right),
 \label{eq:ESN}
\end{align}
where $\mathbf{W}_{\rm rec}$ and $\mathbf{W}_{\rm in}$ are randomly
chosen and fixed parameters\footnote{The spectral radius of
$\mathbf{W}_{\rm rec}$ is set smaller than 1.}, and $\rho$ is a leak
parameter satisfying $0<\rho<1$.  In \eqref{eq:ESN}, tanh is a
hyperbolic tangent function but may be replaced with other nonlinear
functions such as a sigmoid function.

An eligibility trace may be considered as a linear counterpart of the
nonlinear features created by an ESN.  Because these features are
generated by mappings with fixed parameters and just given as input to a
Gaussian DyBM, the learning rules for the Gaussian DyBM stay unchanged.
The nonlinear DyBM in \cite{NonlinearDyBM} uses an ESN in a slightly
different manner.

\subsection{Functional dynamic Boltzmann machines}

We now review a functional DyBM, which models time-series of functions
 (patterns over a continuous space $\mathcal{Z}$) \cite{FunctionalDyBM}.
 Recall that a Gaussian DyBM defines the conditional distribution of the
 next real-valued vector given the preceding sequence of real-valued
 vectors.  A functional DyBM defines the conditional distribution of the
 next function ({\it i.e.}, $g^{[t]}$) given the preceding sequence of
 partial observations of preceding functions.  At each time $s$, a set
 of points $Z^{[s]}\equiv(z_i^{[s]})_{i=1,\ldots,N_s}$ is observed,
 where $N_s$ is the number of points that are observed at $s$.
 
 The functional DyBM assumes that the conditional distribution of
 $g^{[t]}(\cdot)$ is given by a Gaussian process, whose mean
 $\mu^{[t]}(\cdot)$ varies over time depending on preceding functions as
 follows:
\begin{align}
 \mu^{[t]}(z)
 & = b(z) + \sum_{\delta=1}^{d-1} \int_\mathcal{Z} w^{[\delta]}(z,z') \, g^{[t-\delta]}(z') \, {\rm d}z'
 + \sum_{\ell=1}^{L} \int_\mathcal{Z} u_\ell(z,z') \, \alpha_\ell^{[t-1]}(z') \, {\rm d}z'
 \label{eq:fDyBM:mu}
\end{align}
for $x\in\mathcal{Z}$, where $b(\cdot)$ is a functional bias,
$w^{[\delta]}(\cdot,\cdot)$ and $u_\ell(\cdot,\cdot)$ are functional weight for each $\delta$ and for each $\ell$, and
\begin{align}
 \alpha_\ell^{[t-1]}(\cdot)
 & = \sum_{s=-\infty}^{t-d} \lambda_\ell^{t-s-d} \, g^{[s]}(\cdot)
\end{align}
is a functional eligibility trace for each $\ell$.  The covariance
$k_{\sigma^2}(\cdot,\cdot)$ of the Gaussian process consists of two
components such that
 \begin{align}
  k_{\sigma^2}(z,z') = k(z,z') + \sigma^2 \, \delta(z,z'),
 \end{align}
where $k(\cdot,\cdot)$ is a arbitrary kernel, $\delta(\cdot,\cdot)$ is a
delta function, and $\sigma$ is a hyperparameter.

For tractability, Kajino proposes particular parametrization for the functional bias and functional weight \cite{FunctionalDyBM}.
Let $P=(p_1,\ldots,p_M)$ be a set of arbitrarily selected $M$ points in $\mathbf{Z}$
 \begin{align}
b(z) & = k_{\sigma^2}(z,P) \, \mathbf{b} \label{eq:fDyBM:b}\\
w^{[\delta]}(z,z') & = k_{\sigma^2}(z,P) \, \mathbf{W}^{[\delta]} \, k_{\sigma^2}(P,z') \\
u_\ell(z,z') & = k_{\sigma^2}(z,P) \, \mathbf{U}^{[\ell]} \, k_{\sigma^2}(P,z') \label{eq:fDyBM:u}
 \end{align}
 for each $\delta$ and each $\ell$, where
 \begin{align}
 k_{\sigma^2}(z,P) \equiv (k_{\sigma^2}(z,p_i))_{i=1,\ldots,M}
 \end{align}
 is a row vector, and
 \begin{align}
  k_{\sigma^2}(P,z') \equiv (k_{\sigma^2}(p_i,z'))_{i=1,\ldots,M}
 \end{align}
is a column vector.

Because $g^{[t]}$ is in the reproducing kernel Hilbert space with kernel
$k_{\sigma^2}$, substituting \eqref{eq:fDyBM:b}-\eqref{eq:fDyBM:u} into
\eqref{eq:fDyBM:mu} gives the following expression:
\begin{align}
 \mu_\theta^{[t]}(z)
 & = k_{\sigma^2}(z,P) \bigg(
 \mathbf{b}
 + \sum_{\delta=1}^{d-1} \mathbf{W}^{[\delta]} \, g^{[t-\delta]}(P)
 + \sum_{\ell=1}^{L} \mathbf{U}^{[\ell]} \, \alpha_\ell^{[t-1]}(P)\bigg),
 \label{eq:fdybm:mu}
\end{align}
where $g^{[t-\delta]}(P)$ is a column vector with $i$-th element being $g^{[t-\delta]}(p_i)$,
and the eligibility-trace vector $\alpha_\ell^{[t-1]}(P)$ can be recursively updated as follows:
\begin{align}
 \alpha_\ell^{[t]}(P) & = \lambda_\ell \, \big(\alpha_\ell^{[t-1]}(P) + g^{[t-d+1]}(P)\big).
\end{align}
Here, we use
$\theta\equiv(\mathbf{b},\mathbf{W}^{[1]},\ldots,\mathbf{W}^{[d-1]},\mathbf{U}^{[1]},\ldots,\mathbf{U}^{[L]})$
to collectively denote the parameters.

While $g^{[s]}(p_i)$ for $i\in[1,M]$ is not observed, Kajino uses a maximum a posteriori (MAP)
estimator $\hat g^{[s]}(p_i)$ in \cite{FunctionalDyBM}:
\begin{align}
 \hat g^{[s]}(p_i)
 & = \mu_\theta^{[s]}(p_i)
 + k(p_i,Z^{[t]}) \, k_{\sigma^2}(Z^{[t]},Z^{[t]})^{-1} \, \big(g^{[t]}(Z^{[t]}) - \mu_\theta^{[t]}(Z^{[t]})) \big),
 \label{eq:fdybm:map}
\end{align}
where $k_{\sigma^2}(Z^{[t]},Z^{[t]})$ is an $N_s\times N_s$ matrix with
$(i,j)$-th element being $k_{\sigma^2}(z_i^{[t]},z_j^{[t]})$, and
$\mu_\theta^{[t]}(Z^{[t]})$ is a column vector defined analogously to
$g^{[t]}(Z^{[t]})$.

The objective of learning a functional DyBM is to maximize the log likelihood of observed values.
The conditional probability density of the functional values of locations $Z^{[t]}$ at time $t$ is given by
\begin{align}
 p_\theta(g^{[t]}(Z^{[t]}) \mid g^{[<t]})
 \sim \exp\Big( -\frac{1}{2} \big(g^{[t]}(Z^{[t]})-\mu_\theta^{[t]}(Z^{[t]})\big)^\top \, k_{\sigma^2}(Z^{[t]},Z^{[t]})^{-1} \, \big( g^{[t]}(Z^{[t]})-\mu_\theta^{[t]}(Z^{[t]})\big) \Big).
\end{align}
The objective is thus to maximize
 \begin{align}
  f(\theta) & \equiv \sum_t f_t(\theta),
  \intertext{where}
  f_t(\theta) & \equiv \log  p_\theta(g^{[t]}(Z^{[t]}) \mid g^{[<t]}) \\
& =  -\frac{1}{2} \big(g^{[t]}(Z^{[t]})-\mu_\theta^{[t]}(Z^{[t]})\big)^\top \, k_{\sigma^2}(Z^{[t]},Z^{[t]})^{-1} \, \big( g^{[t]}(Z^{[t]})-\mu_\theta^{[t]}(Z^{[t]})\big) + C,
 \end{align}
 where $C$ is the term independent of $\theta$.

 The gradient of $f_t(\theta)$ is given by
 \begin{align}
  \nabla f_t(\theta)
  & = \nabla \mu_\theta^{[t]}(Z^{[t]})^\top \, k_{\sigma^2}(Z^{[t]},Z^{[t]})^{-1} \, \big( g^{[t]}(Z^{[t]})-\mu_\theta^{[t]}(Z^{[t]})\big),
  \intertext{where \eqref{eq:fdybm:mu} gives}
  \frac{\partial}{\partial b_i}  \mu_\theta^{[t]}(Z^{[t]})^\top  & = k_{\sigma^2}(p_i,Z^{[t]}) \\
  \frac{\partial}{\partial w_{i,j}^{[\delta]}} \mu_\theta^{[t]}(Z^{[t]})^\top  & = k_{\sigma^2}(p_i,Z^{[t]}) \, g^{[t-\delta]}(p_j) \\
  \frac{\partial}{\partial u_{i,j}^{[\ell]}} \mu_\theta^{[t]}(Z^{[t]})^\top  & = k_{\sigma^2}(p_i,Z^{[t]})\, \alpha_\ell^{[t-1]}(p_j)
 \end{align}
 for each $i,j,\delta,\ell$.

 The gradient implies the following learning rule with stochastic gradient:
 \begin{align}
  \mathbf{b} & \leftarrow \mathbf{b} + \eta \,
  k_{\sigma^2}(P,Z^{[t]}) \, k_{\sigma^2}(Z^{[t]},Z^{[t]})^{-1} \, \big( g^{[t]}(Z^{[t]})-\mu_\theta^{[t]}(Z^{[t]})\big) \\
  \mathbf{W}^{[\delta]} & \leftarrow \mathbf{W}^{[\delta]} + \eta \,
  k_{\sigma^2}(P,Z^{[t]}) \, k_{\sigma^2}(Z^{[t]},Z^{[t]})^{-1} \, \big( g^{[t]}(Z^{[t]})-\mu_\theta^{[t]}(Z^{[t]})\big) \, g^{[t-\delta]}(P)^\top \\
  \mathbf{U}^{[\ell]} & \leftarrow \mathbf{U}^{[\ell]} + \eta \,
  k_{\sigma^2}(P,Z^{[t]}) \, k_{\sigma^2}(Z^{[t]},Z^{[t]})^{-1} \, \big( g^{[t]}(Z^{[t]})-\mu_\theta^{[t]}(Z^{[t]})\big) \, \alpha_\ell^{[t-1]}(P)^\top,
\end{align}
where $\eta$ is a learning rate, and $g^{[t-\delta]}(P)$ is estimated with the MAP estimator $\hat g^{[t-\delta]}(P)$ in \eqref{eq:fdybm:map}.

\subsection{Dynamic Boltzmann machines with hidden units\protect\footnote{This section closely follows \cite{BidirectionalDyBM}.}}
\label{sec:hidden}

\begin{figure}[t]
 \centering
   \includegraphics[width=0.5\linewidth]{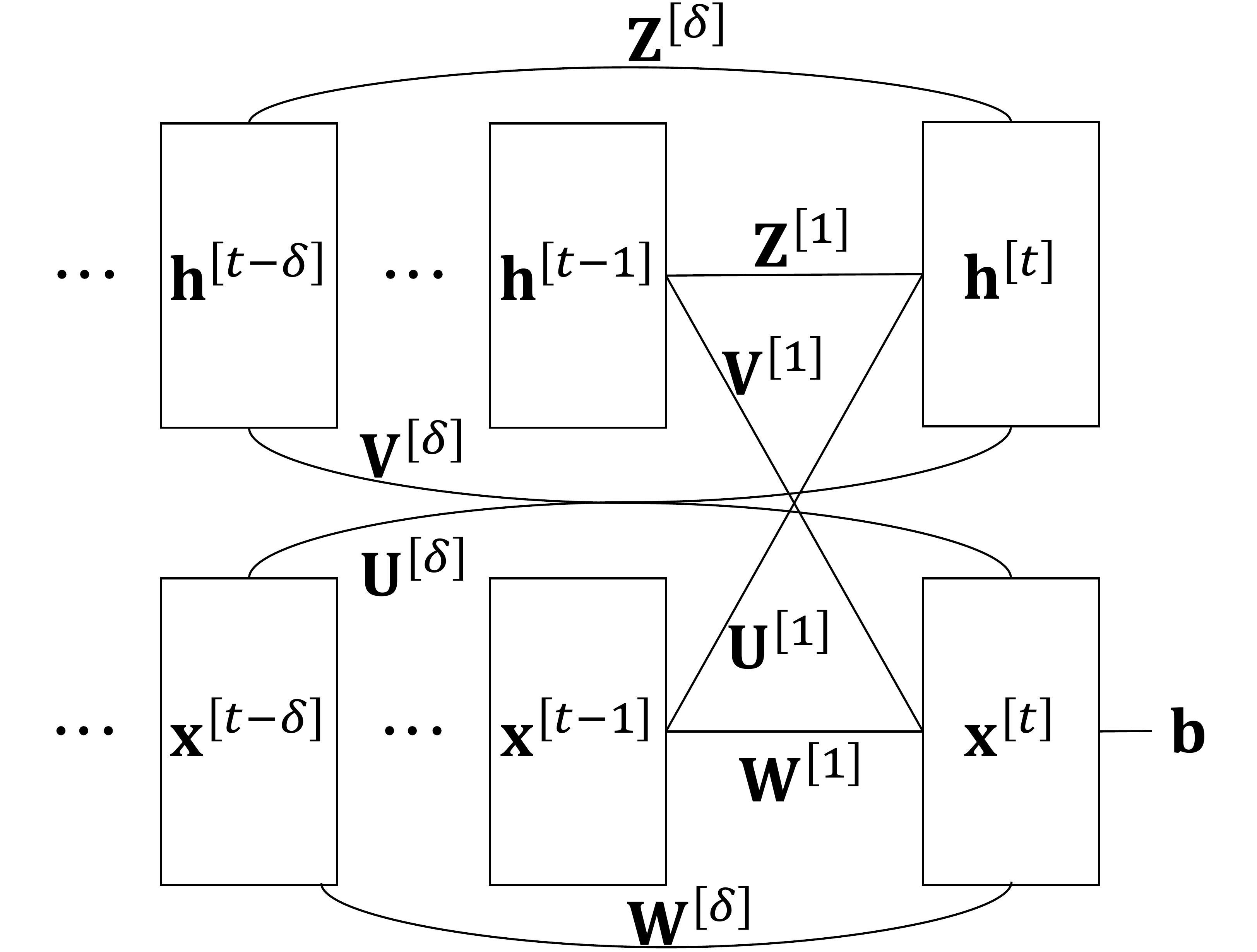}
 \caption{A dynamic Boltzmann machine with hidden units (modified Figure 1 from \cite{BidirectionalDyBM}).}
 \label{fig:DyBM:hidden}
\end{figure}

In this section, we study a DyBM with hidden units (see
Figure~\ref{fig:DyBM:hidden}).  Each layer of this DyBM corresponds to a
time $t-\delta$ for $0\le \delta<\infty$ and has two parts: visible and
hidden. The visible part $\mathbf{x}^{[t-\delta]}$ at the $\delta$-th
layer represents the values of the time-series at time $t-\delta$.  The
hidden part $\mathbf{h}^{[t-\delta]}$ represents the values of hidden
units at time $t-\delta$.  Here, units within each layer do not have
connections to each other.  We let $\mathbf{x}^{[<t]} \equiv
(\mathbf{x}^{[s]})_{s<t}$ and define $\mathbf{h}^{[<t]}$
analogously.

The Boltzmann machine in Figure~\ref{fig:DyBM:hidden} has
bias parameter $\mathbf{b}$ and weight parameter ($\mathbf{U}$,
$\mathbf{V}$, $\mathbf{W}$, $\mathbf{Z}$).
Let $\theta\equiv(\mathbf{V},\mathbf{W},\mathbf{b})$ be the
parameters connected to visible units $\mathbf{x}^{[t]}$ (from the
units in the past, $\mathbf{x}^{[s]}$ or $\mathbf{h}^{[s]}$ for
$s<t$) and $\phi\equiv(\mathbf{U},\mathbf{Z})$.
The conditional energy of this Boltzmann machine is given as follows:
\begin{align}
E_{\theta,\phi}(\mathbf{x}^{[t]},\mathbf{h}^{[t]} \mid \mathbf{x}^{[<t]}, \mathbf{h}^{[<t]})
& = E_\theta(\mathbf{x}^{[t]} \mid \mathbf{x}^{[<t]}, \mathbf{h}^{[<t]})
+E_\phi(\mathbf{h}^{[t]} \mid \mathbf{x}^{[<t]}, \mathbf{h}^{[<t]}),
\label{eq:energy_sum}
\end{align}
where we define
\begin{align}
E_\theta(\mathbf{x}^{[t]} \mid \mathbf{x}^{[<t]}, \mathbf{h}^{[<t]}) 
& = - \mathbf{b}^\top \mathbf{x}^{[t]}
- \sum_{\delta=1}^{\infty} (\mathbf{x}^{[t-\delta]})^\top \mathbf{W}^{[\delta]} \, \mathbf{x}^{[t]}
- \sum_{\delta=1}^{\infty} (\mathbf{h}^{[t-\delta]})^\top \mathbf{V}^{[\delta]} \, \mathbf{x}^{[t]} \\
E_\theta(\mathbf{h}^{[t]} \mid \mathbf{x}^{[<t]}, \mathbf{h}^{[<t]}) 
& = - \mathbf{b}^\top \mathbf{h}^{[t]}
- \sum_{\delta=1}^{\infty} (\mathbf{x}^{[t-\delta]})^\top \mathbf{U}^{[\delta]} \, \mathbf{h}^{[t]}
- \sum_{\delta=1}^{\infty} (\mathbf{h}^{[t-\delta]})^\top \mathbf{Z}^{[\delta]} \, \mathbf{h}^{[t]}.
 \label{eq:energy-finite}
\end{align}
and assume the following parametric form for $\delta\ge d$:
\begin{align}
  \mathbf{W}^{[\delta]} &= \lambda^{\delta-d} \, \mathbf{W}^{[d]}\\
  \mathbf{V}^{[\delta]} &= \lambda^{\delta-d} \, \mathbf{V}^{[d]}\\
  \mathbf{Z}^{[\delta]} &= \lambda^{\delta-d} \, \mathbf{Z}^{[d]}\\
  \mathbf{U}^{[\delta]} &= \lambda^{\delta-d} \, \mathbf{U}^{[d]},
  \label{eq:parametric}
\end{align}
where $\lambda$ is a decay rate satisfying $0\le\lambda<1$.  Then the
conditional energy can be represented as follows:
\begin{align}
\lefteqn{E_\theta(\mathbf{x}^{[t]} \mid \mathbf{x}^{[<t]}, \mathbf{h}^{[<t]})}\notag\\
& = - \mathbf{b}^\top \mathbf{x}^{[t]}
 - \sum_{\delta=1}^{d-1} (\mathbf{x}^{[t-\delta]})^\top \mathbf{W}^{[\delta]} \, \mathbf{x}^{[t]}
 - \sum_{\delta=1}^{d-1} (\mathbf{h}^{[t-\delta]})^\top \mathbf{V}^{[\delta]} \, \mathbf{x}^{[t]} 
  - (\boldsymbol{\alpha}^{[t-1]})^\top \mathbf{W}^{[d]} \, \mathbf{x}^{[t]}
 - (\boldsymbol{\beta}^{[t-1]})^\top \mathbf{V}^{[d]} \, \mathbf{x}^{[t]},
 \label{eq:hidden:energy} \\
\lefteqn{E_\phi(\mathbf{h}^{[s]} \mid \mathbf{x}^{[<s]}, \mathbf{h}^{[<s]})}\notag\\
& = 
 - \sum_{\delta=1}^{d-1} (\mathbf{x}^{[s-\delta]})^\top \mathbf{U}^{[\delta]} \, \mathbf{h}^{[s]}
 - \sum_{\delta=1}^{d-1} (\mathbf{h}^{[s-\delta]})^\top \mathbf{Z}^{[\delta]} \, \mathbf{h}^{[s]}
 - (\boldsymbol{\alpha}^{[s-1]})^\top \, \mathbf{U}^{[d]} \mathbf{h}^{[s]}
 - (\boldsymbol{\beta}^{[s-1]})^\top \, \mathbf{Z}^{[d]} \mathbf{h}^{[s]}.
 \label{eq:energy_hidden}
\end{align}
where $\boldsymbol{\alpha}^{[t-1]}$ corresponds to the eligibility trace
in the DyBM in \eqref{eq:DyBM:etrace}, and we define an eligibility
trace $\boldsymbol{\beta}^{[t-1]}$ for the hidden part analogously:
\begin{align}
 \boldsymbol{\alpha}^{[t-1]}
& \equiv \sum_{\delta=d}^\infty \lambda^{\delta-d} \, \mathbf{x}^{[t-\delta]} \\ 
 \boldsymbol{\beta}^{[t-1]}
& \equiv \sum_{\delta=d}^\infty \lambda^{\delta-d} \, \mathbf{h}^{[t-\delta]}.
\end{align}

The energy in \eqref{eq:hidden:energy}-\eqref{eq:energy_hidden} gives
the conditional probability distribution over $\mathbf{x}^{[t]}$ and
$\mathbf{h}^{[t]}$ given $\mathbf{x}^{[<t]}$ and $\mathbf{h}^{[<t]}$.
Specifically, we have
\begin{align}
\PP_\theta(\mathbf{x}^{[t]} \mid \mathbf{x}^{[<t]}, \mathbf{h}^{[<t]})
 & = \frac{\exp(-E_\theta(\mathbf{x}^{[t]}\mid\mathbf{x}^{[<t]}, \mathbf{h}^{[<t]}))}
 {\displaystyle\sum_{\mathbf{\tilde x}^{[t]}} \exp(-E_\theta(\mathbf{\tilde x}^{[t]}\mid\mathbf{x}^{[<t]}, \mathbf{h}^{[<t]}))} 
 \label{eq:p_theta1} \\
\PP_\phi(\mathbf{h}^{[t]} \mid \mathbf{x}^{[<s]}, \mathbf{h}^{[<s]})
 & = \frac{\exp(-E_\phi(\mathbf{h}^{[t]}\mid\mathbf{x}^{[<s]}, \mathbf{h}^{[<s]}))}
 {\displaystyle\sum_{\mathbf{\tilde h}^{[t]}} \exp(-E_\phi(\mathbf{\tilde h}^{[t]}\mid\mathbf{x}^{[<s]}, \mathbf{h}^{[<s]}))} 
\label{eq:p_phi}
\end{align}
for any binary vectors $\mathbf{x}^{[t]}$ and $\mathbf{h}^{[t]}$.

\subsubsection{Learning a dynamic Boltzmann machine with hidden units}
\label{sec:training}

The DyBM with hidden units gives the probability of a
time-series, $\mathbf{x}\equiv(\mathbf{x}^{[t]})_{t=\ell,\ldots,u}$, by
\begin{align}
 \PP_{\theta,\phi}(\mathbf{x})
 & = \sum_{\mathbf{\tilde h}} \PP_\phi(\mathbf{\tilde h} \mid \mathbf{x})
  \prod_{t=\ell}^{u} \PP_{\theta}(\mathbf{x}^{[t]} \mid \mathbf{x}^{[<t]}, \mathbf{\tilde h}^{[<t]}) 
\label{eq:px}
\end{align}
where $\sum_{\mathbf{\tilde h}}$ denotes the summation over all of the
possible values of hidden units from time $t=\ell$
to $t=u$,
and
\begin{align}
 \PP_\phi(\mathbf{\tilde h} \mid \mathbf{x})
 \equiv
 \prod_{s=\ell}^{u} \PP_\phi(\mathbf{\tilde h}^{[s]} \mid \mathbf{x}^{[<s]}, \mathbf{\tilde h}^{[<s]}),
\label{eq:phx}
\end{align}
where we arbitrarily define
$\mathbf{x}^{[s]}=\mathbf{0}$ and $\mathbf{\tilde h}^{[s]}=\mathbf{0}$
for $s<\ell$.

We seek to maximize the log likelihood of a given $\mathbf{x}$ by
maximizing a lower bound given by Jensen's inequality:
\begin{align}
\log \PP_{\theta,\phi}(\mathbf{x})
 & = \log \Big(
 \sum_{\mathbf{\tilde h}} \PP_\phi(\mathbf{\tilde h} \mid \mathbf{x})
  \prod_{t=\ell}^{u} \PP_{\theta}(\mathbf{x}^{[t]} \mid \mathbf{x}^{[<t]}, \mathbf{\tilde h}^{[<t]}) 
 \Big) \\
 & \ge \sum_{\mathbf{\tilde h}} \PP_\phi(\mathbf{\tilde h} \mid \mathbf{x}) \,
  \log\Big( \prod_{t=\ell}^{u} \PP_{\theta}(\mathbf{x}^{[t]} \mid \mathbf{x}^{[<t]}, \mathbf{\tilde h}^{[<t]}) \Big) \\
 & = \sum_{\mathbf{\tilde h}} \PP_\phi(\mathbf{\tilde h} \mid \mathbf{x}) 
  \sum_{t=\ell}^{u} \log \PP_{\theta}(\mathbf{x}^{[t]} \mid \mathbf{x}^{[<t]}, \mathbf{\tilde h}^{[<t]}) \\
 & = \sum_{t=\ell}^{u} \sum_{\mathbf{\tilde h}^{[<t]}} \PP_\phi(\mathbf{\tilde h}^{[<t]} \mid \mathbf{x}^{[<t-1]}) \,
 \log \PP_{\theta}(\mathbf{x}^{[t]} \mid \mathbf{x}^{[<t]}, \mathbf{\tilde h}^{[<t]}) \notag\\
 & \equiv L_{\theta,\phi}(\mathbf{x}),
 \label{eq:LB}
\end{align}
where the summation with respect to $\mathbf{\tilde h}^{[<t]}$ is over all of the possible values of
$\mathbf{\tilde h}^{[s]}$ for $s\le t-1$, and
\begin{align}
  \PP_\phi(\mathbf{\tilde h}^{[<t]} \mid \mathbf{x}^{[<t-1]})
  \equiv
  \prod_{s=\ell}^{t-1} \PP_\phi(\mathbf{\tilde h}^{[s]} \mid \mathbf{x}^{[<s]}, \mathbf{\tilde h}^{[<s]}).
\end{align}

\paragraph{Learning weight to visible units}

The gradient of the lower bound with respect to $\theta$ is then given by
\begin{align}
\nabla_{\theta} L_{\theta,\phi}(\mathbf{x}) 
 & = \sum_{t=\ell}^{u}
 \sum_{\mathbf{\tilde h}^{[<t]}} \PP_\phi(\mathbf{\tilde h}^{[<t]} \mid \mathbf{x}^{[<t-1]}) \,
 \nabla_{\theta} \log \PP_{\theta}(\mathbf{x}^{[t]} \mid \mathbf{x}^{[<t]}, \mathbf{\tilde h}^{[<t]}).
 \label{eq:grad_theta}
\end{align}
The right-hand side of \eqref{eq:grad_theta} is a 
summation of expected gradients, which suggests a method of stochastic
gradient.  Namely, at each step $t$, we sample $\boldsymbol{H}^{[t-1]}(\omega)$
according to $\PP_\phi(\mathbf{h}^{[t-1]}
\mid \mathbf{x}^{[<t-1]},\mathbf{h}^{[<t-1]})$ and update $\theta$ on the
basis of
\begin{align}
 \nabla_{\theta} \log \PP_{\theta}(\mathbf{x}^{[t]} \mid \mathbf{x}^{[<t]}, \boldsymbol{H}^{[<t]}(\omega)).
\label{eq:grad:visible}
\end{align}

This learning rule is equivalent to the one for the model where all of
the units are visible, except that the values for the hidden units are
given by sampled values.  Therefore, the learning rule for $\theta$
follows directly from Section~\ref{sec:dybm:dybm}:
\begin{align}
 \mathbf{b}
 & \leftarrow \mathbf{b}
 + \eta \, \Big(\mathbf{x}^{[t]} - \E_\theta\big[\boldsymbol{X}^{[t]} \mid \mathbf{x}^{[<t]}, \boldsymbol{H}^{[<t]}(\omega)\big]\Big)
 \label{eq:learning_rule_b} \\
 \mathbf{W}^{[d]} 
 & \leftarrow \mathbf{W}^{[d]}
 + \eta \, \boldsymbol{\alpha}^{[t-1]} \,
 \Big(\mathbf{x}^{[t]} - \E_\theta\big[\boldsymbol{X}^{[t]} \mid \mathbf{x}^{[<t]}, \boldsymbol{H}^{[<t]}(\omega)\big]\Big)^\top
 \label{eq:learning_rule_Wd}\\
 \mathbf{V}^{[d]}
 & \leftarrow \mathbf{V}^{[d]}
 + \eta \, \boldsymbol{\beta}^{[t-1]}(\omega) \,
 \Big(\mathbf{x}^{[t]} - \E_\theta\big[\boldsymbol{X}^{[t]} \mid \mathbf{x}^{[<t]}, \boldsymbol{H}^{[<t]}(\omega)\big]\Big)^\top
 \label{eq:learning_rule_Vd}\\
 \mathbf{W}^{[\delta]}
 & \leftarrow \mathbf{W}^{[\delta]}
 + \eta \, \mathbf{x}^{[t-\delta]} \,
 \Big(\mathbf{x}^{[t]} - \E_\theta\big[\boldsymbol{X}^{[t]} \mid \mathbf{x}^{[<t]}, \boldsymbol{H}^{[<t]}(\omega)\big]\Big)^\top
 \label{eq:learning_rule_W}\\
 \mathbf{V}^{[\delta]}
 & \leftarrow \mathbf{V}^{[\delta]}
 + \eta \, \boldsymbol{H}^{[t-\delta]}(\omega) \,
 \Big(\mathbf{x}^{[t]} - \E_\theta\big[\boldsymbol{X}^{[t]} \mid \mathbf{x}^{[<t]}, \boldsymbol{H}^{[<t]}(\omega)\big]\Big)^\top
 \label{eq:learning_rule_V}
\end{align}
for $1\le\delta<d$, where $\E_\theta[\boldsymbol{X}^{[t]} \mid \mathbf{x}^{[<t]}, \boldsymbol{H}^{[<t]}(\omega)]$ denotes the conditional expectation 
with respect to $\PP_\theta(\cdot \mid \mathbf{x}^{[<t]}, \boldsymbol{H}^{[<t]}(\omega))$, and
we make explicit that $\boldsymbol{\beta}^{[s-1]}$ is computed with sampled hidden values:
\begin{align}
 \boldsymbol{\beta}^{[s-1]}(\omega)
 & = \sum_{\delta=d}^\infty \lambda^{\delta-d} \, \boldsymbol{H}^{[s-\delta]}(\omega).
\end{align}

\paragraph{Learning weight to hidden units}

Now we take the gradient of $L_{\theta,\phi}(\mathbf{x})$ with respect to $\phi$:
\begin{align}
\nabla_{\phi} L_{\theta,\phi}(\mathbf{x}) 
 & = \sum_{t=\ell}^{u} \sum_{\mathbf{\tilde h}^{[<t]}} \nabla_{\phi} p_\phi(\mathbf{\tilde h}^{[<t]} \mid \mathbf{x}^{[<t-1]}) \,
 \log p_{\theta}(\mathbf{x}^{[t]} \mid \mathbf{x}^{[<t]}, \mathbf{\tilde h}^{[<t]}),
 \label{eq:grad_phi1} 
\end{align}
where
\begin{align}
\nabla_{\phi} p_\phi(\mathbf{\tilde h}^{[<t]} \mid \mathbf{x}^{[<t-1]})
  & = \nabla_{\phi} \prod_{s=\ell}^{t-1} p_\phi(\mathbf{\tilde h}^{[s]} \mid \mathbf{x}^{[<s]}, \mathbf{\tilde h}^{[<s]})\\
  & = \sum_{s=\ell}^{t-1}\nabla_{\phi} \log p_\phi(\mathbf{\tilde h}^{[s]} \mid \mathbf{x}^{[<s]}, \mathbf{\tilde h}^{[<s]})
  \prod_{s'=\ell}^{t-1} p_\phi(\mathbf{\tilde h}^{[s']} \mid \mathbf{x}^{[<s']}, \mathbf{\tilde h}^{[<s']})\notag\\
  & = p_\phi(\mathbf{\tilde h}^{[<t]} \mid \mathbf{x}^{[<t-1]})
  \sum_{s=\ell}^{t-1}\nabla_{\phi} \log p_\phi(\mathbf{\tilde h}^{[s]} \mid \mathbf{x}^{[<s]}, \mathbf{\tilde h}^{[<s]}).
\label{eq:grad_phi2}
\end{align}

Plugging \eqref{eq:grad_phi2} into the right-hand side
of \eqref{eq:grad_phi1}, we obtain
\begin{align}
\nabla_{\phi} L_{\theta,\phi}(\mathbf{x})
 & = \sum_{t=\ell}^{u}
  \sum_{\mathbf{\tilde h}^{[<t]}}
  p_\phi(\mathbf{\tilde h}^{[<t]} \mid \mathbf{x}^{[<t-1]}) \,
  \log p_{\theta}(\mathbf{x}^{[t]} \mid \mathbf{x}^{[<t]}, \mathbf{\tilde h}^{[<t]})
  \sum_{s=\ell}^{t-1}\nabla_{\phi} \log p_\phi(\mathbf{\tilde h}^{[s]} \mid \mathbf{x}^{[<s]}, \mathbf{\tilde h}^{[<s]}).
 \label{eq:grad_phi} \end{align}
 Similar to \eqref{eq:grad_theta}, the
 expression of \eqref{eq:grad_phi} suggests a method of stochastic
 gradient: at each time $t$, we sample $\boldsymbol{H}^{[t-1]}(\omega)$ according to
 $p_\phi(\mathbf{h}^{[t-1]}
 \mid \mathbf{x}^{[<t-1]})$ and update
 $\phi$ on the basis of the following stochastic gradient:
\begin{align}
\log p_{\theta}(\mathbf{x}^{[t]} \mid \mathbf{x}^{[<t]}, \boldsymbol{H}^{[<t]}(\omega)) \, G_{t-1},
 \label{eq:grad:hidden}
\end{align}
where
\begin{align}
G_{t-1} \equiv
\sum_{s=\ell}^{t-1} 
\nabla_{\phi} \log p_\phi(\boldsymbol{H}^{[s]}(\omega) \mid \mathbf{x}^{[<s]}, \boldsymbol{H}^{[<s]}(\omega)).
\label{eq:G}
\end{align}

The learning rules for $\mathbf{U}$ and $\mathbf{Z}$ are derived from
\eqref{eq:grad:hidden}-\eqref{eq:G} as follows:
\begin{align}
 \mathbf{U}^{[d]}
 & \leftarrow \mathbf{U}^{[d]}
   + \eta \, \log p_\theta(\mathbf{x}^{[t]}\mid\mathbf{x}^{[<t]},\mathbf{h}^{[<t]})
 \sum_{s=\ell}^{t-1} \boldsymbol{\alpha}^{[s-1]} \,
 \Big(\boldsymbol{H}^{[s]}(\omega) - \E_\phi\big[ \mathbf{H^{[s]}} \mid \mathbf{x}^{[<s]}, \boldsymbol{H}^{[<s]}(\omega)\big]\Big)^\top
     \label{eq:learning_rule_U} \\
 \mathbf{Z}^{[d]}
 & \leftarrow \mathbf{Z}^{[d]}
 + \eta \, \log p_\theta(\mathbf{x}^{[t]}\mid\mathbf{x}^{[<t]},\mathbf{h}^{[<t]})
 \sum_{s=\ell}^{t-1} \boldsymbol{\beta}^{[s-1]}(\omega) \,
 \Big(\boldsymbol{H}^{[s]}(\omega) - \E_\phi\big[ \boldsymbol{H}^{[s]} \mid \mathbf{x}^{[<s]}, \boldsymbol{H}^{[<s]}(\omega)\big]\Big)^\top \\
 \mathbf{U}^{[\delta]}
 & \leftarrow \mathbf{U}^{[\delta]}
   + \eta \, \log p_\theta(\mathbf{x}^{[t]}\mid\mathbf{x}^{[<t]},\mathbf{h}^{[<t]})
 \sum_{s=\ell}^{t-1} \mathbf{x}^{[s-\delta]} \,
 \Big(\boldsymbol{H}^{[s]}(\omega) - \E_\phi\big[ \mathbf{H^{[s]}} \mid \mathbf{x}^{[<s]}, \boldsymbol{H}^{[<s]}(\omega)\big]\Big)^\top \\
 \mathbf{Z}^{[\delta]}
 & \leftarrow \mathbf{Z}^{[\delta]}
 + \eta \, \log p_\theta(\mathbf{x}^{[t]}\mid\mathbf{x}^{[<t]},\mathbf{h}^{[<t]})
 \sum_{s=\ell}^{t-1} \boldsymbol{H}^{[s-\delta]}(\omega) \,
 \Big(\boldsymbol{H}^{[s]}(\omega) - \E_\phi\big[ \boldsymbol{H}^{[s]} \mid \mathbf{x}^{[<s]}, \boldsymbol{H}^{[<s]}(\omega)\big]\Big)^\top
 \label{eq:learning_rule_Z}
\end{align}
for $1\le\delta<d$, where $\E_\phi[\boldsymbol{H}^{[s]}(\omega) \mid \mathbf{x}^{[<s]}, \boldsymbol{H}^{[<s]}(\omega)]$
denotes the conditional expectation with respect to $\PP_\phi(\cdot \mid \mathbf{x}^{[<s]}, \boldsymbol{H}^{[<s]}(\omega))$.

Computation of \eqref{eq:grad_phi} involves mainly two interrelated
inefficiencies.  First, although \eqref{eq:grad_phi} can be
approximately computed using sampled hidden values
$\boldsymbol{H}^{[<t]}(\omega)$ in the same way as \eqref{eq:grad_theta},
the samples cannot be reused after updating $\phi$ because it was
sampled from the distribution with the previous parameter.  Second,
since each summand of $G_{t-1}$ depends on $\phi$, $G_{t-1}$ also has to
be recomputed after each update.  Thus, the computational complexity of
\eqref{eq:grad:hidden} grows linearly with respect to the length of the
time-series ({\it i.e.}, $t-\ell$), in contrast to
\eqref{eq:grad:visible}, whose complexity is independent of that length.

Observe in \eqref{eq:grad:hidden} that $\nabla_{\phi}
L_{\theta,\phi}(\mathbf{x})$ consists of the products of $\log
p_{\theta}(\mathbf{x}^{[t]} \mid \mathbf{x}^{[<t]}, \boldsymbol{H}^{[<t]}(\omega))$
and $\nabla_{\phi} \log p_\phi(\boldsymbol{H}^{[s]}(\omega)
\mid \mathbf{x}^{[<s]}, \boldsymbol{H}^{[<s]}(\omega))$ for $s<t$.  Without the
dependency on $\log p_{\theta}(\mathbf{x}^{[t]}
\mid \mathbf{x}^{[<t]}, \boldsymbol{H}^{[<t]}(\omega))$, the parameter $\phi$ is
updated in a way that $\boldsymbol{H}^{[s]}(\omega)$ is more likely to be
generated ({\it i.e.}, the learning rule would be equivalent to that
for visible units).  Such an update rule is undesirable, because
$\boldsymbol{H}^{[s]}(\omega)$ has been sampled and is not necessarily what we
want to sample again.  The dependency on $\log
p_{\theta}(\mathbf{x}^{[t]} \mid \mathbf{x}^{[<t]}, \boldsymbol{H}^{[<t]}(\omega))$
suggests that $\phi$ is updated by a large amount if the sampled
$\boldsymbol{H}^{[s]}(\omega)$ happens to make the future values,
$\mathbf{x}^{[t]}$ for $t>s$, likely.  Intuitively, weighting
$\nabla_{\phi} \log p_\phi(\boldsymbol{H}^{[s]}(\omega)
\mid \mathbf{x}^{[<s]}, \boldsymbol{H}^{[<s]}(\omega))$ by $\log
p_{\theta}(\mathbf{x}^{[t]} \mid \mathbf{x}^{[<t]}, \boldsymbol{H}^{[<t]}(\omega))$
for $t>s$ is inevitable, because whether the particular values of
hidden units are good for the purpose of predicting future values will
only be known after seeing future values.

\paragraph{Approximations}

One could approximately compute \eqref{eq:G} recursively:
\begin{align}
  G_t
  \leftarrow
  \gamma\,G_{t-1} + (1-\gamma) \nabla_{\phi} \log p_\phi(\boldsymbol{H}^{[t]}(\omega) \mid \mathbf{x}^{[<t]}, \boldsymbol{H}^{[<t]}(\omega)),
  \label{eq:recursive}
\end{align}
where $\gamma\in[0,1)$ is a discount factor.  The recursive update rule
with $\gamma<1$ puts exponentially small weight $\gamma^{t-s}$ on
$\nabla_{\phi} \log p_\phi(\boldsymbol{H}^{[s]}(\omega) \mid
\mathbf{x}^{[<s]}, \boldsymbol{H}^{[<s]}(\omega))$ computed with an old
value of $\phi$ ({\it i.e.}, $s\ll t$).
This recursively computed $G_t$ is related to the momentum in gradient
descent \cite{momentum}.

In \eqref{eq:learning_rule_U}-\eqref{eq:learning_rule_Z}, the value of
$\E_\phi[\boldsymbol{H}^{[s]} \mid \mathbf{x}^{[<s]},
\boldsymbol{H}^{[<s]}(\omega)]$ is computed with the latest values of
$\phi$.  Let $\phi^{[t-1]}$ be the value of $\phi$ immediately before
step $t$.  With the recursive computation of \eqref{eq:recursive}, the
learning rules of \eqref{eq:learning_rule_U}-\eqref{eq:learning_rule_Z}
are approximated with the following learning rules:
\begin{align}
 \mathbf{U}^{[d]}
 & \leftarrow \mathbf{U}^{[d]}
 + \eta \, (1-\gamma) \, \log p_\theta(\mathbf{x}^{[t]}\mid\mathbf{x}^{[<t]},\boldsymbol{H}^{[<t]}(\omega))
 \notag \\ & \hspace{35mm}
 \sum_{s=\ell}^{t-1} \gamma^{t-1-s} \, \boldsymbol{\alpha}^{[s-1]} \,
 \Big(\boldsymbol{H}^{[s]}(\omega) - \E_{\phi^{[s-1]}}\big[ \boldsymbol{H}^{[s]} \mid \mathbf{x}^{[<s]}, \boldsymbol{H}^{[<s]}(\omega)\big]\Big)^\top
 \label{eq:dybm:bi:approx1}\\
 \mathbf{Z}^{[d]}
 & \leftarrow \mathbf{Z}^{[d]}
 + \eta \, (1-\gamma) \, \log p_\theta(\mathbf{x}^{[t]}\mid\mathbf{x}^{[<t]},\boldsymbol{H}^{[<t]}(\omega))
 \notag \\ & \hspace{35mm}
 \sum_{s=\ell}^{t-1} \gamma^{t-1-s} \, \boldsymbol{\beta}^{[s-1]} \,
 \Big(\boldsymbol{H}^{[s]}(\omega) - \E_{\phi^{[s-1]}}\big[ \boldsymbol{H}^{[s]} \mid \mathbf{x}^{[<s]}, \boldsymbol{H}^{[<s]}(\omega)\big]\Big)^\top \\
 \mathbf{U}^{[\delta]}
 & \leftarrow \mathbf{U}^{[\delta]}
   + \eta \, (1-\gamma) \, \log p_\theta(\mathbf{x}^{[t]}\mid\mathbf{x}^{[<t]},\boldsymbol{H}^{[<t]}(\omega))
 \notag \\ & \hspace{35mm}
 \sum_{s=\ell}^{t-1} \gamma^{t-1-s} \, \mathbf{x}^{[s-\delta]} \,
 \Big(\boldsymbol{H}^{[s]}(\omega) - \E_{\phi^{[s-1]}}\big[ \boldsymbol{H}^{[s]} \mid \mathbf{x}^{[<s]}, \boldsymbol{H}^{[<s]}(\omega)\big]\Big)^\top \\
 \mathbf{Z}^{[\delta]}
 & \leftarrow \mathbf{Z}^{[\delta]}
 + \eta \, (1-\gamma) \, \log p_\theta(\mathbf{x}^{[t]}\mid\mathbf{x}^{[<t]},\boldsymbol{H}^{[<t]}(\omega))
 \notag \\ & \hspace{35mm}
 \sum_{s=\ell}^{t-1} \gamma^{t-1-s} \, \boldsymbol{H}^{[s-\delta]}(\omega) \,
 \Big(\boldsymbol{H}^{[s]}(\omega) - \E_{\phi^{[s-1]}}\big[ \boldsymbol{H}^{[s]} \mid \mathbf{x}^{[<s]}, \boldsymbol{H}^{[<s]}(\omega)\big]\Big)^\top
 \label{eq:dybm:bi:approx4}
\end{align}
for $1\le\delta<d$, where $\boldsymbol{H}^{[s]}(\omega)$ is a sample
according to
$\PP_{\phi^{[s-1]}}(\cdot\mid\mathbf{x}^{[<s]},\boldsymbol{H}^{[<s]}(\omega))$
for each $s$.  In \eqref{eq:dybm:bi:approx1}-\eqref{eq:dybm:bi:approx1},
the quantity such as
\begin{align}
 G_{t-1}' \equiv \sum_{s=\ell}^{t-1} \gamma^{t-1-s} \, \boldsymbol{\alpha}^{[s-1]} \,
 \Big(\boldsymbol{H}^{[s]}(\omega) - \E_{\phi^{[s-1]}}\big[ \boldsymbol{H}^{[s]} \mid \mathbf{x}^{[<s]}, \boldsymbol{H}^{[<s]}(\omega)\big]\Big)^\top
\end{align}
can be computed recursively as
\begin{align}
 G_t' \leftarrow \gamma \, G_{t-1}' + (1-\gamma) \, \boldsymbol{\alpha}^{[t-1]} \,
 \Big(\boldsymbol{H}^{[t]}(\omega) - \E_{\phi^{[s-1]}}\big[ \boldsymbol{H}^{[t]} \mid \mathbf{x}^{[<t]}, \boldsymbol{H}^{[<t]}(\omega)\big]\Big)^\top.
\end{align}

In \cite{BidirectionalDyBM}, we present an alternative approach of
learning the DyBM with hidden units in a bidirectional manner, where we
consider a backward DyBM that shares the parameters of the (forward)
DyBM.  Our key observation is that the parameters that are difficult to
learn in the forward DyBM are relatively easy to learn in the backward
DyBM.  By training both the forward DyBM and the backward DyBM, we can
effectively learn the parameters of the forward DyBM.

\section{Conclusion}

We have reviewed Boltzmann machines for time-series modeling.  Such
Boltzmann machines can be used for prediction
\cite{NonlinearDyBM,BidirectionalDyBM,FunctionalDyBM,RTRBM,RNNRBM,LSTM-RTRBM} and
anomaly detection based on observed time-series.  They may be also used
to generate time-series such as human motion \cite{CRBM,TayHin09,RTRBM}, music \cite{LSTM-RTRBM}, and movies.

The use of Boltzmann machines is only one approach to modeling and
learning time-series.  Popular time-series models include but not
limited to recurrent neural networks \cite{RNN}, long short term memory
\cite{LSTM}, autoregressive models, and hidden Markov models.  As we
have seen some of the examples, the best time-series model for a
particular application might be obtained by appropriately combining some
of existing time-series models.

\section*{Acknowledgments}
This work was supported by JST CREST Grant Number JPMJCR1304, Japan.  The author thanks Diyuan Lu for pointing out several typographical errors in the original version.

\bibliographystyle{plain}
\bibliography{dybm,gradient,neuro,rnn,boltzmann,timeseries,dpp,energy,RL}

\end{document}